\documentclass{article}

\PassOptionsToPackage{numbers}{natbib}

\usepackage[preprint]{neurips_2024}

\usepackage[utf8]{inputenc} 
\usepackage[T1]{fontenc}    
\usepackage{hyperref}       
\usepackage{url}            
\usepackage{booktabs}       
\usepackage{amsfonts}       
\usepackage{nicefrac}       
\usepackage{microtype}      
\usepackage[dvipsnames,table,xcdraw]{xcolor}                    
\usepackage{multirow}       
\usepackage[createShortEnv]{proof-at-the-end}
\usepackage{amsthm}
\usepackage{thmtools,thm-restate}
\newtheorem{theorem}{Theorem}[section]
\newtheorem{corollary}{Corollary}[theorem]

\newtheorem{prop}{Proposition}
\usepackage{pythonhighlight}
\usepackage{wrapfig}        
\usepackage{svg}
\usepackage{amsmath}        
\usepackage{bm}
\usepackage{xr-hyper}
\usepackage{sidecap,wasysym,array,tabularx,caption}

\usepackage{caption}
\usepackage{xcolor}

\definecolor{Gray}{gray}{0.9}
\usepackage{amssymb}
\usepackage{amsmath}

\usepackage{appendix}
\newcommand*{\Comb}[2]{{}^{#1}C_{#2}}%

\usepackage[at]{easylist}

\title{FouRA: \underline{Fou}rier Low \underline{R}ank \underline{A}daptation}

\author{%
  Shubhankar Borse\thanks{These authors contributed equally to this work.} \quad Shreya Kadambi$^*$ \quad Nilesh Prasad Pandey \quad Kartikeya Bhardwaj \\ 
  \textbf{Viswanath Ganapathy} \quad \textbf{Sweta Priyadarshi} \quad \textbf{Risheek Garrepalli} \quad \textbf{Rafael Esteves} \\ \textbf{Munawar Hayat} \quad \textbf{Fatih Porikli}\\ 
  Qualcomm AI Research\thanks{Qualcomm AI Research is an initiative of Qualcomm Technologies, Inc.}\\
  \texttt{\{sborse, skadambi, nileshpr, kbhardwa, viswgana, swetpriy, rgarrepa, resteves,}\\ 
  \texttt{mhayat, fporikli\}@qti.qualcomm.com} \\
}
\begin{document}

\maketitle

\begin{abstract}

While Low-Rank Adaptation (LoRA) has proven beneficial for efficiently fine-tuning large models, LoRA fine-tuned text-to-image diffusion models lack diversity in the generated images, as the model tends to copy data from the observed training samples. This effect becomes more pronounced at higher values of adapter strength and for adapters with higher ranks which are fine-tuned on smaller datasets. To address these challenges, we present FouRA, a novel low-rank method that learns projections in the Fourier domain along with learning a flexible input-dependent adapter rank selection strategy. Through extensive experiments and analysis, we show that FouRA successfully solves the problems related to data copying and distribution collapse while significantly improving the generated image quality. We demonstrate that FouRA enhances the generalization of fine-tuned models thanks to its adaptive rank selection. We further show that the learned projections in the frequency domain are decorrelated and prove effective when merging multiple adapters. While FouRA is motivated for vision tasks, we also demonstrate its merits for language tasks on the GLUE benchmark.

  
\end{abstract}

\vspace{- 1.0 em}
\section{Introduction}
\label{sec:introduction}




\begin{figure*}[h]
    \centering
    \includegraphics[width=1.0\linewidth]{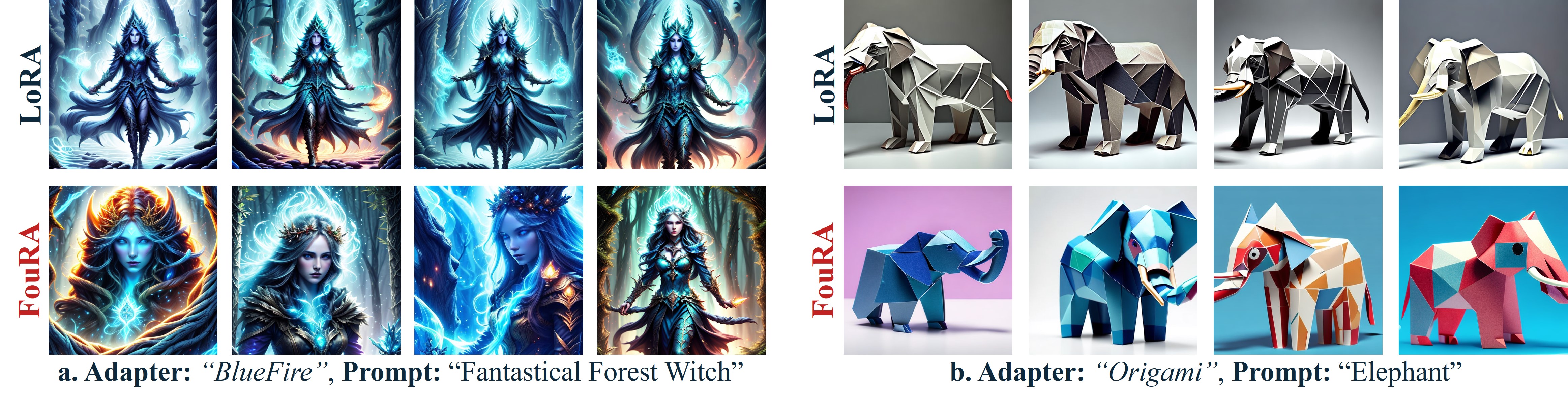}
    \caption{\textbf{Distribution collapse with LoRA}. Visual results generated by the Realistic Vision 3.0 model trained with LoRA and FouRA, for "\textit{Blue Fire}" and "\textit{Origami}" style adapters \textbf{across four seeds}. While LoRA images suffer from distribution collapse and lack diversity, we observe diverse images generated by FouRA.}
    \label{fig:mode_collapse}
    \vspace{- 0.5 em}
\end{figure*}

Parameter-Efficient FineTuning (PEFT)~\cite{peft} methods such as Low-Rank Adaptation~\cite{lora} provide a promising solution to quickly adapt large foundation models, including large vision models (LVMs) and large language models (LLMs) to new tasks~\cite{dora, rslora, ding2023sparse}. 
The LoRA module has an elegant design, allowing quick adaptation to new styles or concepts without changing the underlying base model, thus effectively retaining previous knowledge and preventing catastrophic forgetting. 

While LoRAs are highly effective in quickly adapt to new styles, they exhibit multiple challenges, with the rank of LoRA modules being a highly sensitive parameter. As LoRA is built for adapting to new tasks using a small training set, it tends to overfit to the  distribution of small training set when the rank is high. Recent works~\cite{somepallidiffusion, somepalli2023understanding} observed that when diffusion models overfit to a small training set, they demonstrate a tendency to repeatedly "copy" few samples from the training set. LoRAs trained on smaller data therefore tend to generate data copying artifacts, also known as distribution collapse. The generated images lack diversity, and the phenomenon is very similar to mode collapse observed in GANs. We illustrate this tendency in Fig.~\ref{fig:mode_collapse}, specially at high values of adapter strength $\alpha$ across different seeds.
Additionally, as the rank reduces, the strength of the adapter reduces, and LoRA has a reduced ability to generate diverse images due to underfitting. Hence, the rank is a very sensitive parameter. 

Gating mechanisms have been proposed~\cite{ding2023sparse} to produce a dynamic rank at every layer, to provide flexibility to the adapter in LLM tasks. However, we argue that dynamic rank reduction is still not flexible for vision tasks as the rank is computed during training and does not vary at inference. We observe that text-to-image diffusion models greatly benefit from a rank adaptation mechanism which can also vary during inference, along the diffusion time steps.
Furthermore, while all the previous works apply low-rank adaptation in the feature space, we argue that there is a transform domain over which fine-tuning low-rank adaptation modules generates much richer representations. 
We provide theoretical and analytical evidence to show that low-rank adaptation in the frequency domain produces a highly compact representation, effectively reducing the generalization error. Hence, this can potentially push the adaptive rank selection mechanism to generalize better, not only reducing the risk of underfitting when rank reduces, but also overfitting at higher ranks. 
Additionally, there have been attempts to merge multiple LoRA concepts and/or styles as a linear weighted combination of multiple LoRAs~\cite{Ryu_Low-rank_adaptation_for}. Recent works~\cite{wu2023mole, mix-of-show, kumari2023multi} empirically show that this approach is prone to noisy and inaccurate outputs, and propose \textbf{joint} finetuning the adapters with learnable gates in the low rank subspace. However, we argue that jointly training multiple LoRA modules is highly restrictive and equally tedious for practical use-cases requiring flexibility in combining multiple different LoRAs. Our developed approach of gating in frequency domain enables flexible mixing of multiple adapters.

In this paper, we propose FouRA (Fourier Low Rank Adaptation), a PEFT technique to address the aforementioned challenges of LoRA. We transform the input features to the frequency domain, and apply both the down-projection (to a lower rank) and the up-projection (back to the higher rank) in this frequency domain. During inference, we fold the adapter strength $\alpha$ into the low rank subspace. FouRA learns an adaptive mask inside the low-rank subspace to dynamically drop certain frequency transformed basis, effectively varying the rank for each layer. The adaptive mask selection is input dependant, and varies during the diffusion process. Through rigorous analysis, we show that FouRA provides clear benefits over LoRA (and other adaptive gating methods), and generates high quality diverse images.We show for lower ranks increasing the effect of adapter weights in FouRA does not deteriorate the representation power of original model. Additionally, we show that FouRA provides a rich disentangled orthogonal basis to Low Rank Adapters in the frequency domain, making it beneficial for merging multiple styles. Our contributions are summarized as:
\begin{itemize}\setlength{\itemsep}{-0.2em}
    \item We introduce FouRA, the first low-rank adapter module that performs the low rank transforms in the frequency domain along pixel or channel dimensions of the feature space. 
    \item We propose an adaptive learnable masking strategy in the frequency domain that flexibly varies the effective rank for every FouRA layer in the network, thus enabling the model to generalize well, even when the size of training set is very small.
    \item We demonstrate that FouRA successfully provides a decorrelated orthonormal basis to Low Rank Adapters in the frequency domain, making it highly beneficial for merging two styles or concepts, without the need for joint training.
    \item Through extensive experiments and theoretical analysis, we demonstrate how FouRA consistently produces a diverse set of aesthetically improved images compared to LoRA, and is equally effective for LLM tasks.
\end{itemize}

\section{Related Work}
\label{sec:Related_Work}

\noindent \textbf{Text-to-Image Diffusion Models:}
Multiple diffusion based image generative models have been proposed recently \cite{sd,sdxl,SDv3}, \cite{dalle,nichol2022glide,ImageGen,wurstchen}. These models exhibit excellent text-to-image generation ability and can be adapted to new styles using LoRA \cite{lora}. 









\noindent \textbf{Fourier Transforms in Generative Literature:}
Recent work \cite{kadkhodaie2023generalization} shows that the latents of the denosing models trained on sufficient data lie on adaptive basis with oscillating patterns. 
Other works have shown that we can use fourier operators for non parametric regression tasks and cast self attention as a kernel regression problem. \cite{nguyen2022transformer} shows that it offers smoother representations over the input and better captures the correlations between query and keys. \cite{li2020fourier} has shown that Fourier spectral filters operate in the continuous domain and work well in representing images as continuous functions. Further convolutions in spatial domain can be represented as multiplications in the Fourier space thus spectral filters can act as global convolution operator. Applying these transforms to Low-Rank space has not been explored before, to the best of our knowledge. 

Many works have analysed the eigen spread of signal transformed to harmonic basis. \cite{beaufays1993simple}, analysed the effect of applying these transforms on a signal sampled from a  Markovian process and show that Fourier transforms decorrelates such as signal in least mean square setting.  

\noindent \textbf{Low Rank Adaptation:}
LoRAs \cite{lora} suffer from a tradeoff between fidelity and diversity of generated images. \cite{ding2023sparse} tried to alleviate this problem by sparse regularization.  SVDiff \cite{han2023svdiff} explicitly only updates the singular values while retaining the subspaces. 
In a high rank setting this method is acceptable. However, in  FouRA we are learning in a low rank subspace. Other works like AdaLORA \cite{zhang2023adaptive}, \cite{xu2023autolora} applied to language models, further parameterized the weight matrices using SVD and jointly optimized for eigen vectors and the singular values through importance scoring metric. O-lora \cite{wang2023orthogonal} computes orthogonal gradient spaces between different tasks letting the model sequentially adapt to new tasks without catastrophic forgetting. 
\cite{ding2023sparse} applies proximal gradient gating in the loss function to learn important subspaces and mask out the remaining ones. 
While all these papers directly operate by constraining the subspace of the weight matrices, we show in our paper that the Fourier domain implicitly enforces these properties without any constraints in the optimization. We show that applying gating in the frequency domain  provides a more compact representation with stable generalization error bounds. In addition results in lower effective rank for each layer. We also show that the learnt spaces across different adapters also have decorrelated basis. MoLE~\cite{wu2023mole}, ZipLoRA\cite{shah2023ziplora} and Mix of Show \cite{mix-of-show, zhong2024multi} explore various strategies to merge LoRAs. This is done using either supervised or self-supervised objectives for jointly training weights corresponding to both adapters. As the number of adapters grow, we argue that the two-stage method to merge adapters is not flexible and quite tedious. FouRA on the other hand does not require any fine-tuning, and is truly a training-free approach to merge multiple adapters.

\noindent \textbf{Disentangled spaces for editing} %
 \cite{wu2023uncovering} \cite{haas2023discovering} have explored diffusion models for disentangled interpretable latent representation. While LoRAs have been proposed for personalization,~\cite{gandikota2023concept} proposed a way to do fine-grained editing of images while still preserving the features of the original image. They identify semantic directions and traverse on the latent space on these directions. Concept sliders have been applied to real applications such as fixing  distortions in diffusion generated images. We show in our work that our method identifies more compact disentangled representations over LoRA, thus providing more performance improvements over fine-gradined edits.
 



\vspace{- 0.5 em}
\section{Proposed Approach}
\label{sec:method}
\vspace{- 0.5 em}


\subsection{Formulation of Low Rank Adaptation}
\label{sec:lora_form}
\begin{figure*}[t]
    \centering
    \includegraphics[width=1.0\linewidth]{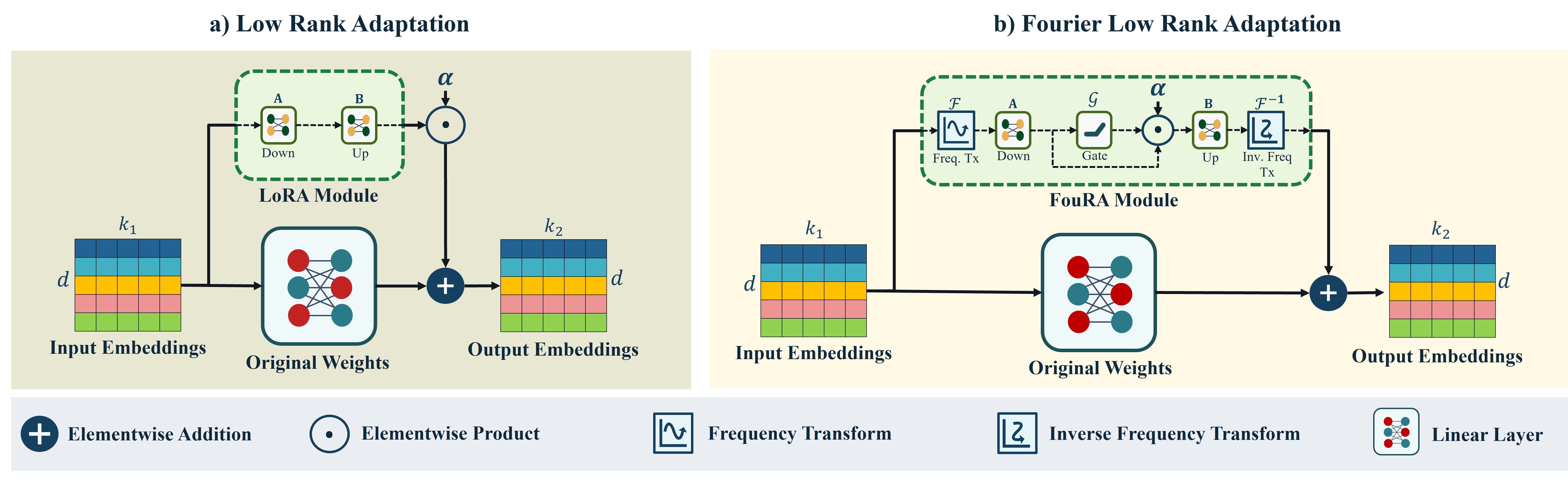}
    \caption{\textbf{LoRA v/s FouRA}. For FouRA, we transform feature maps to frequency domain, where we learn up and down adapter projections along-with our proposed adaptive rank gating module.}
    \label{fig:linear_foura}
    \vspace{- 1.5 em}
\end{figure*}

We illustrate the base LoRA module in Fig.~\ref{fig:linear_foura}. Consider the original set of pre-trained weights $\mathbf{W_0} \in \mathbb{R}^{k_1 \times k_2}$ where $k_1$ and $k_2$ represent the input and output embedding dimensions respectively. LoRA modules consist of the down layer $\mathbf{A} \in \mathbb{R}^{k_1 \times r}$ and the up layer $\mathbf{B} \in \mathbb{R}^{r \times k_2}$, projecting the input features to and from the low-rank subspace of rank $r$. 
Consider an input feature $\mathbf{z_{in}} \in \mathbb{R}^{d \times k_1}$, where $d$ is the number of input tokens, the output after the low-rank adaptation $\mathbf{z_{out}} \in \mathbb{R}^{d \times k_2}$ is given as $\mathbf{z_{out}} = \mathbf{z_{og}} + \alpha\mathbf{z_{lora}} = \mathbf{W_0} \mathbf{z_{in}} + \alpha\mathbf{B}\mathbf{A}\mathbf{z_{in}}$. Here, $\mathbf{z_{og}}$ and $\mathbf{z_{lora}}$ are the outputs from the original and low-rank branches respectively, and $\alpha$ is a scalar to blend the two branches. We denote the learned adapter matrices as $\Delta\mathbf{W_{lora}} = \mathbf{B}\mathbf{A}$ as in~\cite{lora}.

\subsection{Low Rank Adaptation in the Frequency Domain}
The projection to and from a low-rank subspace is prone to information loss. To mitigate this, we propose to transform the inputs to a domain which contains an inherently compact representation, i.e. the frequency domain. We are motivated by the fact that transforming to the frequency domain preserves valuable information, due to its inherent de-correlation capabilities~\cite{2016ChellappaDCT, he2024frequencyadaptive}. We validate this further by analyzing the effects of the frequency transform on the model weights in Sec.~\ref{sec:theory_freqdomain}. 

Given the pre-trained weight matrix $\mathbf{W_0}$, we apply the low rank transforms $\mathbf{B}$ and $\mathbf{A}$ in the frequency domain. Inspired by~\cite{si2023freeu}, we fold the blending parameter $\alpha$ inside the low-rank subspace, effectively acting as a scaling factor in the frequency domain. We apply the frequency transforms as follows.

\vspace{- 1.0 em}
\begin{equation}
\label{eq:freq}
    \mathbf{z_{out}} = \mathbf{z_{og}} + \mathbf{z_{foura}} = \mathbf{W_0} \mathbf{z_{in}} + \mathcal{F}^{-1}(\mathbf{B}\alpha\mathbf{A}\mathcal{F}(\mathbf{z_{in}}))
\end{equation}


Here, $\mathcal{F}(\cdot)$ and $\mathcal{F}^{-1}(\cdot)$ are the normalized forward and inverse frequency transforms respectively. 


\subsection{Frequency Transforms}
We investigate the properties of Discrete Fourier Transform (DFT) and Discrete Cosine Transform (DCT) in the low rank space.  We apply 1D DFT to the embedding dimension $k_{1} \in (0,K)$ before the subspace decomposition. Given input $z_{in} \in \mathbb{R}^{d \times k_1}$ to the adapter branch , we expand  $\mathcal{F}$ in Eq. \eqref{eq:freq} as, 
\begin{equation}
    \label{eq:fft}
     \mathbf{Z_{k_{1}}}(f)= \mathcal{F}(\mathbf{z_{in}})_{d \times k_{1}} =  \frac{1}{\sqrt{k_{1}}}\sum_{k=0}^{k_{1}-1} e^{-j \frac{2\pi f_r k}{k_{1}}} \mathbf{z_{in}}(k), f_{r} : \forall r \in (0, 1...k_{1}-1).
\end{equation}

Where $f_r$ is the frequency of the basis represented by DFT. As we do not apply any padding, the dimension of the transform preserves the dimension of $\mathbf{z_{in}}$. In our experiments, we apply the 1-D transform on the embedding dimension $k_{1}$ for each token on both self and cross attention layers. 

To motivate the idea of generalizing FouRA across tasks such as targeted editing \cite{gandikota2023concept}, where disentangled latent space is required to gain control over generated images, we further explored Discrete Cosine Transform (DCT) with compact subspaces (eigen spread), which leads to less overfitting. We later show in App.~\ref{sec:appendix_svd} and Fig.~\ref{fig:sv_distribution} that the subspaces of FouRA are more uncorrelated from each other. 
We observe that for certain tasks, DCT provides a smoother representation as the implicit window is twice that of DFT signals. 
For a given a finite length signal $\mathbf{z_{in}} \in \mathrm{R}^{d \times k_{1}} $, we compute DCT as follows. We first construct a double length even signal by 
  \begin{equation}
    \mathbf{\tilde{z_{in}}}(d, k_{1})=
    \begin{cases}
      \mathbf{z_{in}}(d, k_{1}), &  0 \leq k_{1} \leq K \\
      \mathbf{z_{in}}(d, 2K-k_{1} -1), &  K \leq k_{1} \leq 2K-1, \\
    \end{cases}
  \end{equation}
The DCT is then computed as the DFT of $\mathbf{\tilde{z_{in}}}$.

\subsection{Adaptive Rank Gating Method}
LoRA methods pre-define the rank for all layers. Recent method~\cite{ding2023sparse} has an adaptive rank during training, which is however fixed at inference time, thus lacking flexibility. In our approach, we propose a learned adaptive gating mechanism, which can vary each layers rank during training and inference, dependent upon the inputs. We introduce our learnable gating mechanism $\mathcal{G}(\cdot)$ inside the low-rank subspace within the frequency domain. 
Consider the low-rank representation denoted as $\mathbf{z_{lr}} \leftarrow \mathbf{A}\mathcal{F}(\mathbf{z_{in}})  \in \mathbb{R}^{d \times r}$, our gating operation is defined as,

\vspace{- 1.5 em}    
\begin{equation}
\label{eq:gate}
    \mathcal{G}(\mathbf{z_{lr}}) = 
    \begin{cases}
        1, & \text{if} \; \mathcal{S}(\mathcal{H}(\mathbf{G}\mathbf{z_{lr}}))==1 \\
        0, & \text{otherwise}
    \end{cases}
\end{equation}
\vspace{- 1.0 em}

\begin{figure}[t]
    \centering
    \includegraphics[width=0.9\linewidth]{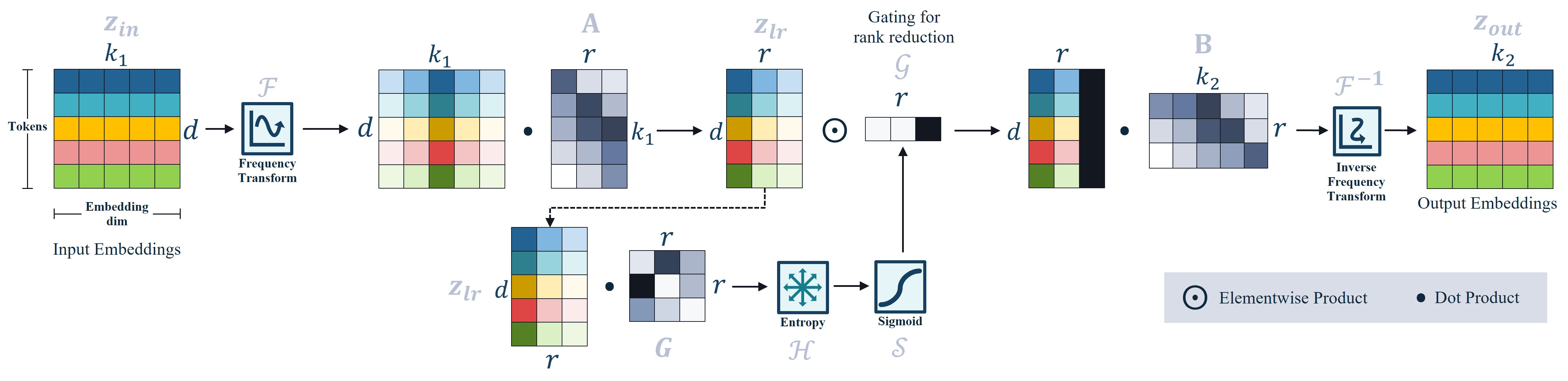}
    \caption{\textbf{Operational diagram of FouRA}. Illustrating the components of Eq.~\ref{eq:freq}.}
    \label{fig:operations}
    \vspace{- 2 em}
\end{figure}

Here, $\mathcal{H}(\cdot)$ and $\mathcal{S}(\cdot)$ represent entropy and sigmoid functions respectively, $\mathbf{G}$ represents the weights of a learnable multi-layer perceptron (MLP), $\mathcal{G}$ is a function to learn a weighting for every singular value in the low-rank subspace. The FouRA output, illustrated in Fig.~\ref{fig:operations}, is then given by,

\vspace{- 1.2 em}
\begin{equation}
\label{eq:freq}
    \mathbf{z_{out}} = \mathbf{z_{og}} + \mathbf{z_{foura}} = \mathbf{W_0} \mathbf{z_{in}} + \mathcal{F}^{-1}(\mathbf{B}\alpha\mathcal{G}(\mathbf{z_{lr}})\cdot\mathbf{A}\mathcal{F}(\mathbf{z_{in}}))
\end{equation}
\vspace{- 1.2 em}

The learned FouRA adapter weights are $\Delta\mathbf{W_{foura}} = \mathcal{F}^{-1}(\mathbf{B}\mathcal{G}(\mathbf{z_{lr}})\mathcal{F(\mathbf{A})})$, as per notation in Sec.~\ref{sec:lora_form}.

We conduct further analysis of our proposed gating function in Sec.~\ref{sec:theory_gating}, analyzing its behaviour across diffusion time-steps and various resolutions. Further, we demonstrate its efficacy over both fixed LoRA and recent Adaptive Rank selection methods which are fixed at inference (SoRA~\cite{ding2023sparse}). 

\subsection{Combining multiple adapters}
Merging of LoRA adapters has multiple practical use-cases~\cite{Ryu_Low-rank_adaptation_for}. The method we use to merge two adapters varies according to the task.

\textbf{Text-to-Image Style Transfer:} Following the standard method, we merge two FouRA style based adapters using a linear combination of the outputs of adapter $\Delta W_{1}. \mathbf{z_{in}}$ and $\Delta W_{2}. \mathbf{z_{in}}$ during inference.

\textbf{Image editing using Concept Sliders:} Similar to \cite{gandikota2023concept}, we perform concept slider evaluations for text based editing using FouRA in Sec.~\ref{sec:concept_slider}. Given $n$ concept sliders,  we define $c_{n,j}$  concept for $n^{th}$ slider (e.g "very old")  and  $\tilde{c}_{n,i}$  as the  negative concept (e.g "  very young"). We composite the adapters in the epsilon $\epsilon$ space, with composed score function $\hat{\epsilon}$, and sample from the factorized distribution $p(\mathbf{\mathbf{x}}/(\tilde c_{i}, c_{j}))$

\vspace{- 1.5 em}
\begin{equation}
    \hat{\epsilon}(\mathbf{x}) = \epsilon_{\theta}(\mathbf{x}) + \sum_{n} w_{n}(\epsilon_{\theta}(\mathbf{x}, c_{n,j}) - \epsilon_{\theta}(\mathbf{x}, c_{n,i}))
\end{equation}
\vspace{- 1.2 em}

For merging of two styles, as well as composition of two concept adapters across different strengths $\alpha$, we notice that the feature spaces of FouRA adapters are less entangled as compared to LoRA. Further analysis is present in Appendix~\ref{sec:appendix_disentangled} and~\ref{sec:appendix_sparsity}. 


\section{Theoretical Analysis}
\label{sec:method}


\subsection{Frequency Domain Fine Tuning}
\label{sec:theory_freqdomain}
Frequency domain transforms decorrelate input representations, minimize spectral redundancy~\cite{zhang2020DCT}, and are effective in compression since they concentrate most of the energy in a few coefficients~\cite{he2024frequencyadaptive}. Learning in the spectral domain is shown to enable faster convergence and sparser weight matrices~\cite{2016ChellappaDCT}. Motivated by these advantages, we propose to fine-tune adapters in the frequency domain.

\begin{wrapfigure}{r}{0.28\textwidth}
    \centering \includegraphics[width=0.26\textwidth]{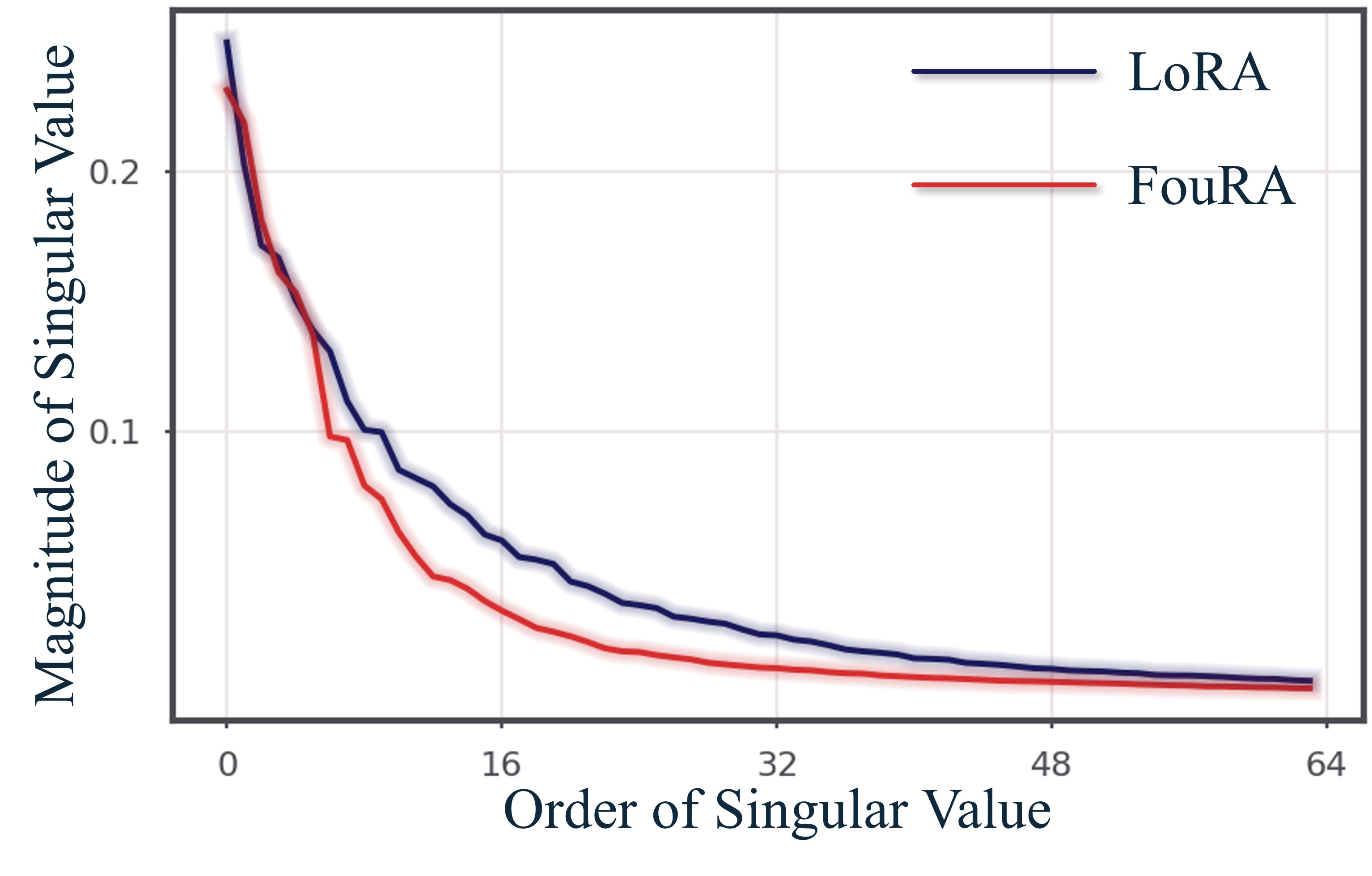}
    \caption{Singular value spread for FouRA v/s LoRA.}
    \label{fig:sv_distribution}
    \vspace{- 1 em}
\end{wrapfigure}

\paragraph{Singular Value Distribution Analysis:}Consider a weight matrix $\mathbf{W}$. The singular value decomposition of this matrix is represented as $\mathbf{U}\mathbf{D}\mathbf{V^T}$, where $\mathbf{U} \in \mathbb{R}^{k_1 \times k_1}, \mathbf{V} \in \mathbb{R}^{k_2 \times k_2}$ are orthonormal matrices and $\mathbf{D} \in \mathbb{R}^{k_1 \times k_2}$ is a matrix, containing the singular values of $\mathbf{W}$, $\sigma_i \forall i\in \{\mathbb{N}^{min(k_1, k_2)}\}$. Considering an $r$ rank approximation of $\mathbf{W}$, we denote the singular values as $\{\sigma_1, \sigma_2 ... \sigma_r\}$, arranged in descending order, and the corresponding diagonal matrix as $\mathbf{D_r}$. The $r$-rank approximation of $\mathbf{W}$ is hence computed as $LR_r(\mathbf{W}) = \mathbf{U}\mathbf{D_r}\mathbf{V^T}$.

\begin{lemmaE}
\label{lem:svd}
Considering two adapters $\Delta\mathbf{W_1}$ and $\Delta\mathbf{W_2}$ and their corresponding sets of singular values $\{\sigma_{1,i}\}$ and $\{\sigma_{2,i}\}$. The adapter $\Delta\mathbf{W_1}$, will admit $r$ rank approximation with lower error than $\Delta\mathbf{W_2}$ if $\sigma_{1,i} < \sigma_{2,i}$ for all $i \geq r$.
\end{lemmaE}

\setlength{\intextsep}{0pt}%
\setlength{\columnsep}{2pt}%

We provide a proof for the above lemma in Appendix~\ref{sec:appendix_svd}. We empirically analyze the distribution of singular values for $r$ rank approximations of $\Delta\mathbf{W_{lora}}$ and $\Delta\mathbf{W_{foura}}$ for the last layer of our trained UNet model in Fig.~\ref{fig:sv_distribution}. FouRA has a more compact spread of singular values as compared to LoRA. Hence, using Lemma~\ref{lem:svd}, we can say that the accumulated error for a LoRA adapter with a low-rank approximation will be greater than the a FouRA adapter with the same rank.

\begin{figure*}[t]
    \centering
    \includegraphics[width=0.9\linewidth]{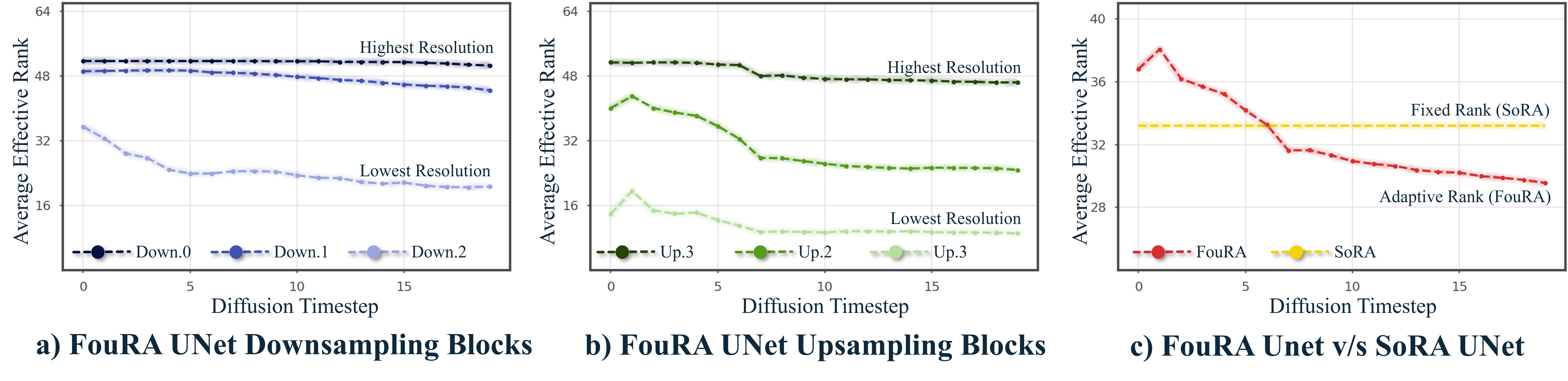}
    
    \caption{\textbf{Average Effective Rank of FouRA}. Figure a. and b. shows plots for the average effective rank for various layers of the FouRA U-Net (Darker lines correspond to higher resolutions) and Figure c. compares the average effective rank of FouRA to SoRA. FouRA's effective rank reduces with the feature resolution, and it also reduces as the diffusion process proceeds, owing to lesser changes required towards the end.}
    \label{fig:effective_rank_graphs}
    \vspace{- 1.2 em}
\end{figure*}

\subsection{Gated Frequency Domain Fine Tuning}
\label{sec:theory_gating}

Motivated by observations in~\cite{ding2023sparse, lin2024lora}, our proposed rank gating mechanism intends to vary the effective rank of each low-rank adapter in the network. We describe effective rank per layer as the number of singular values which are not masked out by the learned gating function. Using observations from~\cite{fu2023effectiveness, lin2024lora}, we propose the following Lemma:

\begin{lemmaE}
\label{lem:sparsity}
Consider an adapter $\Delta\mathbf{W}$ with a rank higher than the required rank to fit a training data distribution. The upper-bound of generalization error $\mathcal{R}$ for fine-tuning this adapter reduces as the effective rank of the adapter reduces. After reducing to a certain value of effective rank, the upper-bound of generalization error will increase as rank reduces further.
\end{lemmaE}

\begin{corollary}
\label{lem:f_sparsity}
 Additionally, the generalization bound is more stable when the singular value distribution of adapter weights $\Delta\mathbf{W}$ is more compact.
\end{corollary}

We provide a proof in Appendix~\ref{sec:appendix_sparsity}. The effectiveness of variable rank selection can be justified using Lemma~\ref{lem:sparsity}. As LoRA rank reduces, the model tends to underfit. However, increasing the rank above the required rank to fit a training distribution leads to overfitting, which reduces the models performance. Dynamically determining the effective rank in every layer produces promising results, as it provides a learnable trade-off between generalization and overfitting. 

In Fig.~\ref{fig:effective_rank_graphs}, we plot FouRA average effective ranks for a denoising UNet over 20 iterations of the reverse diffusion process. Our analysis reveals that the effective rank learnt for high-resolution layers is higher than low-resolution layers. Furthermore, the effective rank reduces as the denoising process continues. This essentially means that noisy inputs require more singular values to update.
We further observe in Fig.~\ref{fig:adarank_selection} that our proposed adaptive masking (which varies in inference time) significantly outperforms methods such as SoRA (which freezes its masks after training).

Furthermore, from Corollary~\ref{lem:f_sparsity} and a consequence of the property observed in Fig.~\ref{fig:sv_distribution}, as FouRA obtains compact spread of singular values, we can determine that the generalization bound is more stable in the frequency domain for lower effective ranks, as compared to the feature space. We verify this in Fig.~\ref{fig:adarank_selection} as FouRA outperforms SoRA and LoRA with our proposed adaptive masking. The \textbf{data copying artifacts} observed for LoRA model in Fig.~\ref{fig:mode_collapse} are a consequence of overfitting. This was observed by recent works targeting Digital Forgery~\cite{somepallidiffusion, somepalli2023understanding}. As FouRA significantly reduces the generalization error, it can generate a diverse set of images. Additionally, we also observe in App.~\ref{sec:unseen} that FouRA is able to generalize better on unseen concepts, as compared to LoRA.

\vspace{- 1.2 em}

\subsection{Subspace Learning}
In App.~\ref{appendix_theory}, we provide a subspace perspective to verify empirically and theoretically that FouRA learns subspaces which are more decorrelated from the base model weights, as compared to LoRA. A higher emphasis on the set of learnt subsapces enables FouRA to learn new tasks without catastrophic forgetting. Additionally, we attribute the strong merging capabilities of different FouRA adapters to their disentangled and decorrelated subspaces learned by respecitve FouRAs.



\section{Experiments}
\label{sec:experiments}

\begin{figure*}[t]
    \centering
    \includegraphics[width=1.0\linewidth]{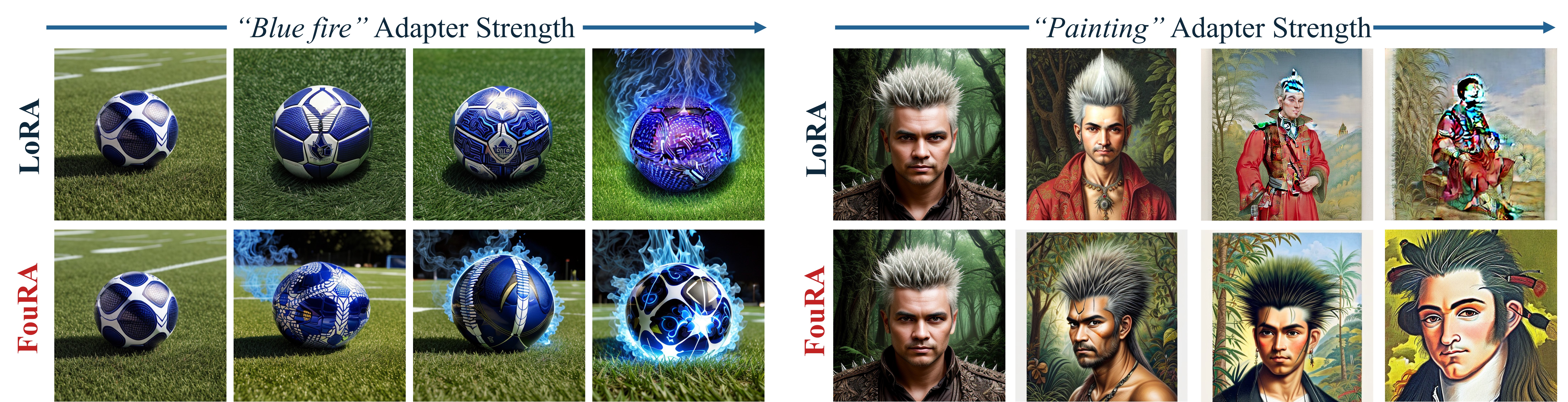}
    \caption{\textbf{FouRA v/s LoRA:} The prompt on the left is "a football in a field" and on the right is "man in a mythical forest". While staying more faithful to the adapter style, FouRA outputs look aesthetically better than LoRA, which have obvious artifacts at high values of $\alpha$. Additional results are in Appendix~\ref{sec:app_t2i}.}
    \label{fig:foura_base}
    \vspace{- 1 em}
\end{figure*}

\subsection{Experimental setup}
\noindent \textbf{Datasets:}  For style transfer, we evaluate FouRA on four datasets collected from public domains, including \textit{Bluefire} , \textit{Paintings}, \textit{3D} and \textit{Origami} styles, see Appendix~\ref{style-transfer-details} for details. 
Our results are averaged over 30 random seeds, and a total of 1530 images. 
For evaluations on composite sliders, similar to \cite{gandikota2023concept}, we train 3 sliders \textit{"Age"}, "\textit{Hair}"  \textit{"Surprised'}  and composite experiments  combining both  \textit{"Age"} and "\textit{Hair}"  .
While our approach is motivated for vision tasks, we also evaluate FouRA on language tasks and assess the performance of our adapter on MNLI, CoLA, SST2, STSB, MRPC and QNLI tasks from the GLUE benchmarks, see App.~\ref{language-dataset} for details.

\noindent \textbf{Models:} For text-to-image generation experiments, we employ Stable Diffusion-v1.5~\cite{sd}, using both the base model weights and RealisticVision-v3.0 checkpoints for style transfer tasks. For concept editing, we train on Stable Diffusion-v1.5~\cite{sd} base weights. We use RoBERTA-Base~\cite{liu2019roberta} for General Language Understanging tasks. See App.~\ref{appendix_impl} for additional implementation details.

\noindent \textbf{Metrics:} For quantifying the quality of images generated by FouRA and LoRA finetuned diffusion models, we report HPSv2.1~\cite{wu2023human} and LPIPS diversity \cite{zhang2018perceptual} scores. The HPSv2 metric evaluates the measure of the image quality, and alignment with the prompt/style. LPIPS diversity score captures the diversity within all possible pairs of generated images across seeds. We provide an in-depth analysis of these metrics in Appendix~\ref{sec:metricinterpret}. For the image editing task, we compare edited images using LPIPS similarity (compared to the base image). For language models, we report on the General Language Understanding Evaluation (GLUE) benchmarks~\cite{wang2018glue}, see details in App.~\ref{tab: gleu}.

        



\subsection{Text-to-Image Stylized Generation}

In Fig.~\ref{fig:foura_base}, we show visual results of LoRA and FouRA on the \textit{Paintings} and \textit{Bluefire} style tasks. FouRA is able to generate high quality images as compared to LoRA over a range of adapter strengths $\alpha$. We observe that LoRA suffers from artifacts at high values of $\alpha$ in case of the Paintings adapter.  
Tab.~\ref{tab: t2i_table} compares LPIPS Diversity and HPSv2 scores for all models, showing that FouRA significantly outperforms LoRA on both the metrics. Our analysis in App.~\ref{sec:metricinterpret} shows that this gap in LPIPS-diversity and HPS scores is quite significant, specially for higher $\alpha$ values, FouRA shows significant gains compared to LoRA. This is likely because at lower $\alpha$ values, the adapter effect would be reduced and thus both images look more realistic. These results demonstrate that FouRA images are both diverse (even at high adapter strengths) as well as aesthetically coherent. See App.~\ref{sec:app_t2i} for more experiments.

\setlength{\intextsep}{0pt}%
\setlength{\columnsep}{5pt}%
\begin{wraptable}{r}{0.31\textwidth}
\hspace{1.0 em}
\centering
    \addtolength{\tabcolsep}{-1pt}
    \centering
    \fontsize{7.0pt}{5.75pt}\selectfont
    \begin{tabular} {  l|cc|c }
    \toprule
    \textbf{Adapter} & $\mathbf{\alpha_b}$ & $\mathbf{\alpha_p}$ & \textbf{HPSv2 score} \\
    \hline
    \hline
    LoRA & 0.4 & 0.4 & 33.4 \\ 
    \rowcolor{Gray} FouRA & 0.4 & 0.4 & \textbf{33.5} \\ 
    \hline
    LoRA & 0.6 & 0.6 & 32.7 \\ 
    \rowcolor{Gray} FouRA & 0.6 & 0.6 & \textbf{33.5} \\ 
    \hline
    LoRA & 0.8 & 0.8 & 31.2 \\ 
    \rowcolor{Gray} FouRA & 0.8 & 0.8 & \textbf{33.6} \\ 
    \hline

    LoRA & 1.0 & 1.0 & 30.3 \\ 
    \rowcolor{Gray} FouRA & 1.0 & 1.0 & \textbf{33.1} \\ 
    \bottomrule
    \end{tabular}
    \vspace{-1 em}
    \caption{Merging two adapters for Blue Fire and Paintings with strengths $\alpha_b$ and $\alpha_p$.} 
    \label{tab:merge_styles}
\end{wraptable}

\begin{figure*}[t]
    \centering
    \includegraphics[width=1.0\linewidth]{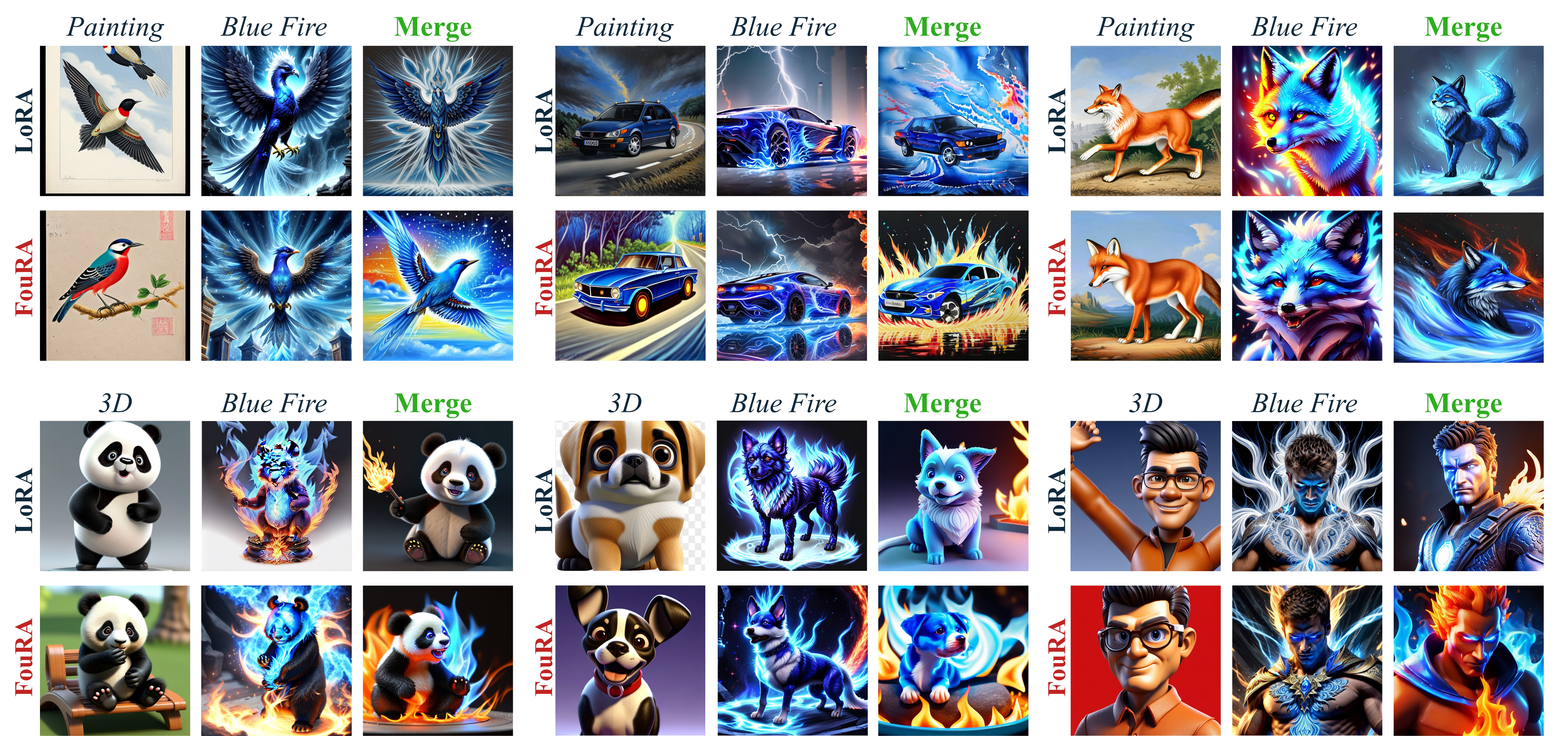}
    \caption{\textbf{Multi-Adapter Fusion in LoRA v/s FouRA}. Sample images for style transfer on various prompts (e.g., bird, car, fox) for \textit{Paintings}, \textit{Bluefire}, \textit{3D }and Merged adapters. Observe the highlighted \textcolor{ForestGreen}{merged} images. FouRA does a much better job in preserving both styles, compared to LoRA.}
    \label{fig:adapter_merging}
    \vspace{- 1 em}
\end{figure*}

\begin{table*}[t]
    \addtolength{\tabcolsep}{-1.5pt}
    \centering
    \scalebox{0.9}{
    \fontsize{7.0pt}{5.75pt}\selectfont
    \begin{tabular} {lc|c|ccc|ccc}
    \toprule
    
    \multirow{2}{*}{\textbf{Dataset}} & \multirow{2}{*}{\textbf{Base Model}} & \multirow{2}{*}{\textbf{Adapter}} & \multicolumn{3}{c|}{\textbf{LPIPS Diversity($\uparrow$)}} & \multicolumn{3}{c}{\textbf{HPSv2 score($\uparrow$)}} \\

    & & & $\alpha=1$ & $\alpha=0.8$ & $\alpha=0.6$ & $\alpha=1$ & $\alpha=0.8$ & $\alpha=0.6$ \\
    \hline
    \midrule
    
    \multirow{4}{*}{\parbox{1.3cm}{Paintings (630 Images)}} & \multirow{2}{*}{Stable Diffusion-v1.5} & LoRA & $38.3 \pm 3.6$ & $43.0 \pm 3.2$ & $43.6 \pm 3.6$ & $22.3 \pm 1.7$ & $25.3 \pm 1.9$ & $27.2 \pm 2.9$ \\
    & &\cellcolor{Gray} FouRA & \cellcolor{Gray} $\bm{43.9 \pm 3.7}$ & \cellcolor{Gray} $\bm{44.1 \pm 3.8}$ & \cellcolor{Gray} $\bm{45.7 \pm 3.8}$ & \cellcolor{Gray} $\bm{25.2 \pm 1.6}$ & \cellcolor{Gray} $\bm{27.1 \pm 1.8}$ & \cellcolor{Gray} $\bm{28.0 \pm 2.4}$ \\
    \cmidrule{2-9}
    & \multirow{2}{*}{Realistic Vision-v3.0} & LoRA & $38.3 \pm 3.5$ & $37.8 \pm 3.6$ & $39.2 \pm 3.7$ & $24.6 \pm 1.8$ & $27.7 \pm 1.8$ & $30.3 \pm 1.7$ \\
    & & \cellcolor{Gray} FouRA & \cellcolor{Gray} $\bm{44.2 \pm 3.7}$ & \cellcolor{Gray} $\bm{44.5 \pm 4.0}$ & \cellcolor{Gray} $\bm{44.6 \pm 3.9}$ & \cellcolor{Gray} $\bm{28.4 \pm 1.8}$ &\cellcolor{Gray} $\bm{30.6 \pm 1.5}$ &\cellcolor{Gray} $\bm{32.0 \pm 1.4}$  \\
    \midrule
    \multirow{4}{*}{\parbox{1.3cm}{Blue-Fire (900 Images)}} & \multirow{2}{*}{Stable Diffusion-v1.5} & LoRA & $47.8 \pm 3.7$ & $48.4 \pm 3.9$ & $49.5 \pm 4.2$ & $28.6 \pm 2.1$ & $30.4 \pm 2.0$ & $30.6 \pm 2.2$ \\
    & & \cellcolor{Gray} FouRA & \cellcolor{Gray} $\bm{50.3 \pm 3.0}$ & \cellcolor{Gray} $\bm{50.8 \pm 3.2}$ & \cellcolor{Gray} $\bm{51.5 \pm 3.6}$ & \cellcolor{Gray}  $\bm{29.7 \pm 1.9}$ & \cellcolor{Gray} $\bm{30.9 \pm 1.9}$ & \cellcolor{Gray} $\bm{30.9 \pm 2.2}$  \\
    \cmidrule{2-9}
    & \multirow{2}{*}{Realistic Vision-v3.0} & LoRA & $46.8 \pm 4.0$ & $48.5 \pm 4.0$ & $49.8 \pm 4.2$ & $32.7 \pm 1.6$ & $33.8 \pm 1.4$ & $34.0 \pm 1.5$ \\
    & & \cellcolor{Gray} FouRA & \cellcolor{Gray} $\bm{50.4 \pm 3.0}$ & \cellcolor{Gray} $\bm{51.6 \pm 3.3}$ & \cellcolor{Gray} $\bm{52.2 \pm 3.5}$ & \cellcolor{Gray} $\bm{33.6 \pm 1.5}$ & \cellcolor{Gray} $\bm{34.1 \pm 1.2}$ & \cellcolor{Gray} $\bm{34.0 \pm 1.4}$ \\
    
    \bottomrule
\end{tabular}
    }
    \caption{Evaluation of LoRAs on Text-to-Image tasks. Adapters are rank 64. Results are averaged over 30 seeds.} 
    \label{tab: t2i_table}
\end{table*}
\textbf{Multi-Adapter:} Fig.~\ref{fig:adapter_merging} shows images for style transfer merging for various prompts (e.g., bird, car, fox) for three styles: \textit{Paintings}, \textit{Bluefire} and \textit{3D}. We also provide the outputs of the linear combination of LoRA and FouRA for both these tasks. We see that merged LoRA images sometimes lose one of the concepts (e.g., the blue fire is lost for Panda and Dog) or have severe artifacts (e.g., the fox with multiple tails and the bird without a head). In comparison, FouRA images for merged adapters preserve the concepts and do not display any distortions. This property of FouRA is a direct consequence of our analysis in App.~\ref{sec:decorr} and is also evident from the HPSv2 reported in Tab.~\ref{tab:merge_styles}, where for higher adapter strengths, FouRA shows gains upto 3\% over LoRA.

\subsection{Text-to-Image Concept Editing}\label{sec:concept_slider}
\begin{figure*}[t]
    \centering
    \includegraphics[width=0.95\linewidth]{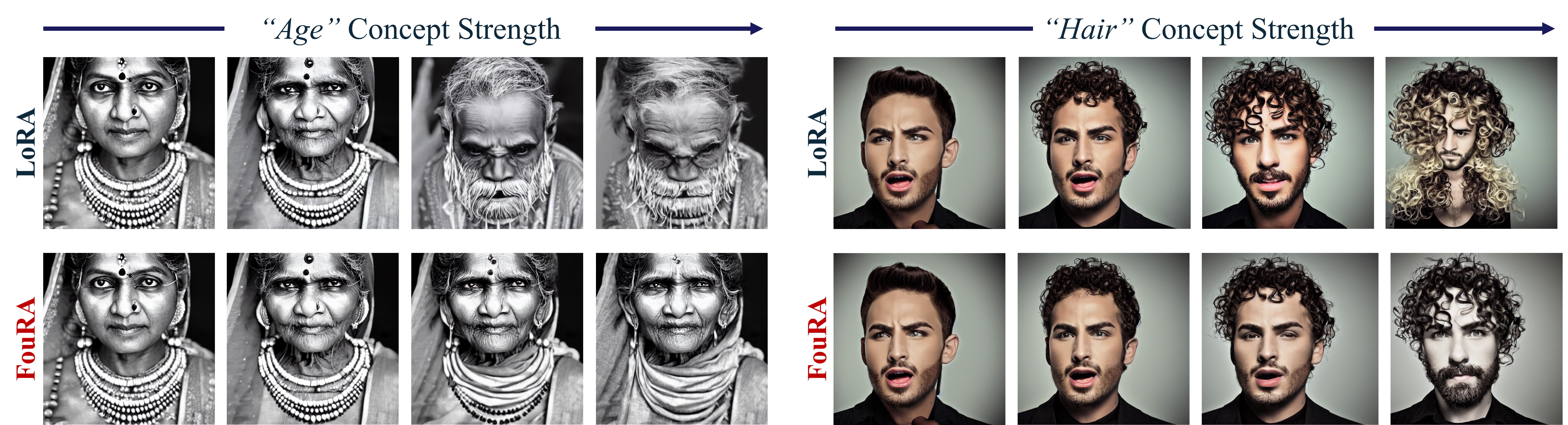}
    \caption{\textbf{LoRA v/s FouRA }. \textit{Age} (Left) and \textit{Hair }(right) concept slider examples where as the scale increases the effect of disentanglement in FouRA is more prominent. For larger scales the gender of the person changes in \textit{Age} LoRA, and the structure of the face changes in \textit{Hair} LoRA.}
    \label{fig:age_slider_1}
    \vspace{- 1 em}
\end{figure*}

We establish the performance of our approach on nuanced editing tasks for specific target images by training FouRA using the disentangled objective proposed in concept sliders ~\cite{gandikota2023concept}. We train LoRA and FouRA modules using pairs of prompts describing the editing concepts. 
Fig.~\ref{fig:age_slider_1} shows results of editing the \textit{Age} and \textit{Hair} concepts. As observed, although the Age adapters are trained using a disentangled objective, LoRA changes the gender of the subject, and produces artifacts at high scales. FouRA is elegantly able to age them while retaining their original features. Similarly, the \textit{Hair} FouRA produces a smoother representation. We provide quantitative evaluations in App.~\ref{sec:concept_slider}, and observe that at higher strengths, FouRA consistently outperforms LoRA in terms of the LPIPS score. 



\noindent\textbf{Composite Sliders:} We qualitatively evaluate the composite  \textit{'hair'} and \textit{'age'} adapter between LoRA and FouRA in Appendix~\ref{sec:concept_slider}. We show the results on two target prompts "A female Indian person" and " A male white person" respectively.  Overall, we observe that FouRA does a better job at compositing both sliders, as it produces a smooth transition between the concepts. In comparison, LoRA distorts the subjects faces at high adapter scales, and interferes with other facial features. We also show that the LPIPS diversity is much lower for generated images between different strength for FouRA \ref{fig:lpips_sliders} at higher scales of the adapter.

\subsection{General Language Understanding Tasks}

While our design choices for FouRA are primarily motivated for vision tasks, we evaluate its efficacy on langauge tasks in Tab.~\ref{tab:glue_adarank}, and compare FouRA against another adaptive rank selection approach, SoRA, designed specifically for language tasks~\cite{ding2023sparse}. Results show that FouRA's rank selection in frequency domain outperforms SoRA on four out of the six GLUE benchmarks we evaluated on, demonstrating that the feature disentanglement induced by FouRA can be used beyond vision tasks.

\vspace{.3 em}
\begin{SCtable}[][h]\setlength{\tabcolsep}{5pt}
\centering
\scalebox{0.75}{

    \begin{tabular} {  l|cccccc }
    \toprule
    \textbf{Adapter} & \textbf{MNLI} & \textbf{CoLA} & \textbf{SST2} & \textbf{STSB} & \textbf{MRPC} &  \textbf{QNLI} \\\hline
    \midrule
    LoRA & $90.2 \pm 0.2$ & $67.3 \pm 0.8$ & $94.9 \pm 0.3$ & $89.9 \pm 0.3$ & $90.3 \pm 0.6$ & $93.6 \pm 0.6$  \\
    SoRA & $\bm{90.5 \pm 0.1}$ & $69.9 \pm 0.8$ & $95.2 \pm 0.4$ & $91.4 \pm 0.1$ & $\bm{90.6 \pm 0.8}$ & $93.9 \pm 0.3$ \\
    \rowcolor{Gray} FouRA & $\bm{90.5 \pm 0.1}$ & $\bm{70.6 \pm 0.7}$ & $\bm{95.5 \pm 0.4}$ & $\bm{91.6 \pm 0.1}$ & $90.4\pm 0.5$ & $\bm{94.2 \pm 0.5}$ \\
    \bottomrule
\end{tabular}
}
    \caption{Evaluation of RoBERTa Models on the GLUE benchmarks, averaged over 3 seeds.} 
    \label{tab:glue_adarank}
    \end{SCtable}
    
\vspace{- 1.2 em}
\label{sec:ablation}

\subsection{Ablation Studies}

\noindent \textbf{Varying the Adaptive Rank Selection Strategy in Text-to-Image Stylized Generation:} 

\setlength{\intextsep}{0pt}%
\setlength{\columnsep}{5pt}%
\begin{wrapfigure}{r}{0.3\textwidth}
    \centering
\includegraphics[width=0.28\textwidth]{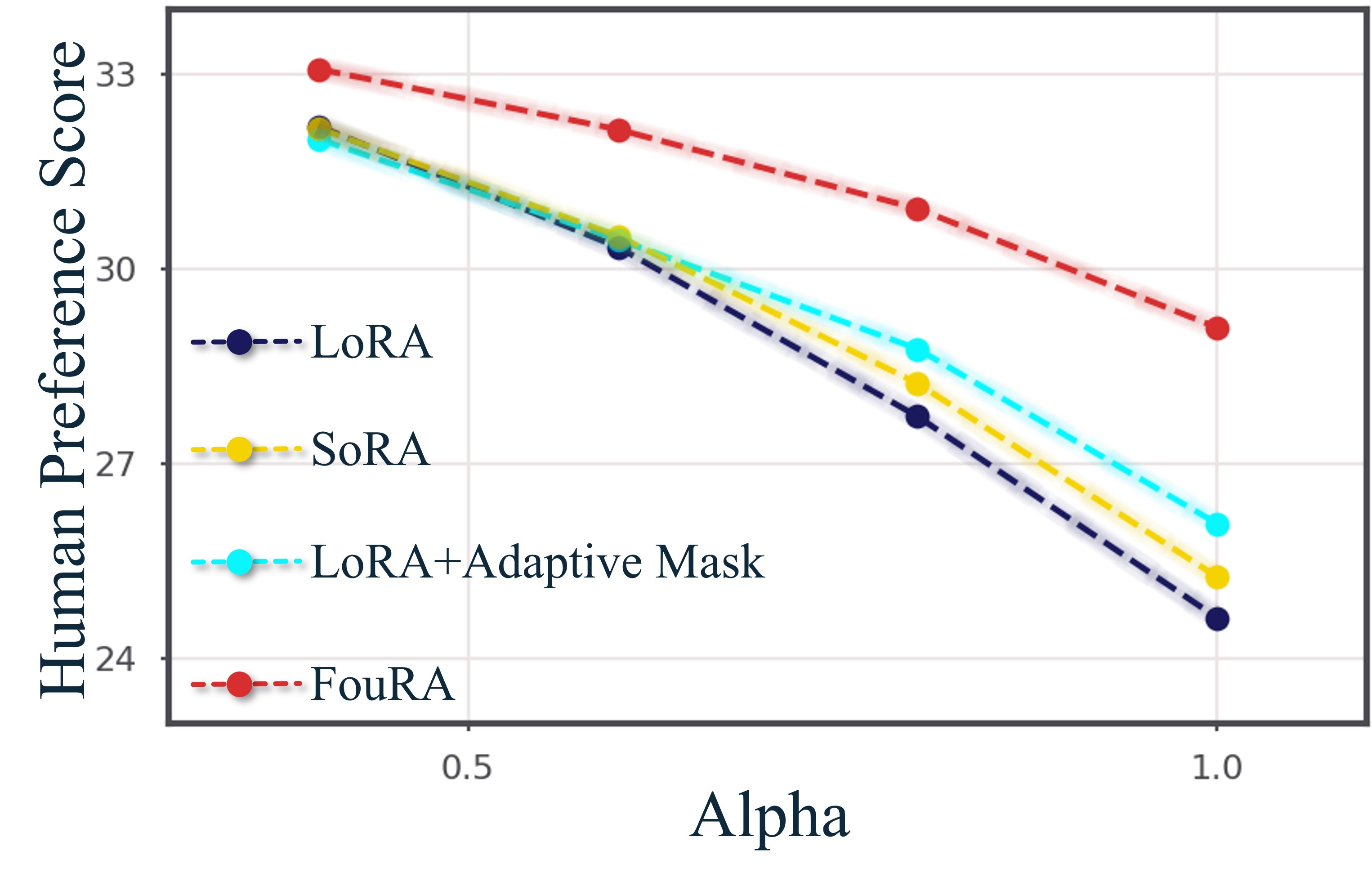}
    \vspace{- 0.7 em}
    \caption{Comparison of different rank selection methods.}
    \label{fig:adarank_selection}
\end{wrapfigure}
Fig.~\ref{fig:adarank_selection} shows the HPS-v2.1 curves obtained for evaluating LoRA, SoRA~\cite{ding2023sparse} and FouRA on the Paintings validation set for different adapter strength $\alpha$. Additionally, we also show the performance of our inference-adaptive rank selection method directly on LoRA. All the numbers are for base rank=64 adapters. As observed, SoRA outperforms LoRA at higher ranks. However, our inference-adaptive rank selection strategy improves performance over SoRA, indicating that in vision models, varying the effective-rank across time steps of the diffusion process is ideal. FouRA outperforms all methods, indicating the benefits of training our proposed rank selection strategy in the frequency domain.

\begin{figure*}[h]
    \centering
    \vspace{1.0 em}
    \includegraphics[width=0.9\linewidth]{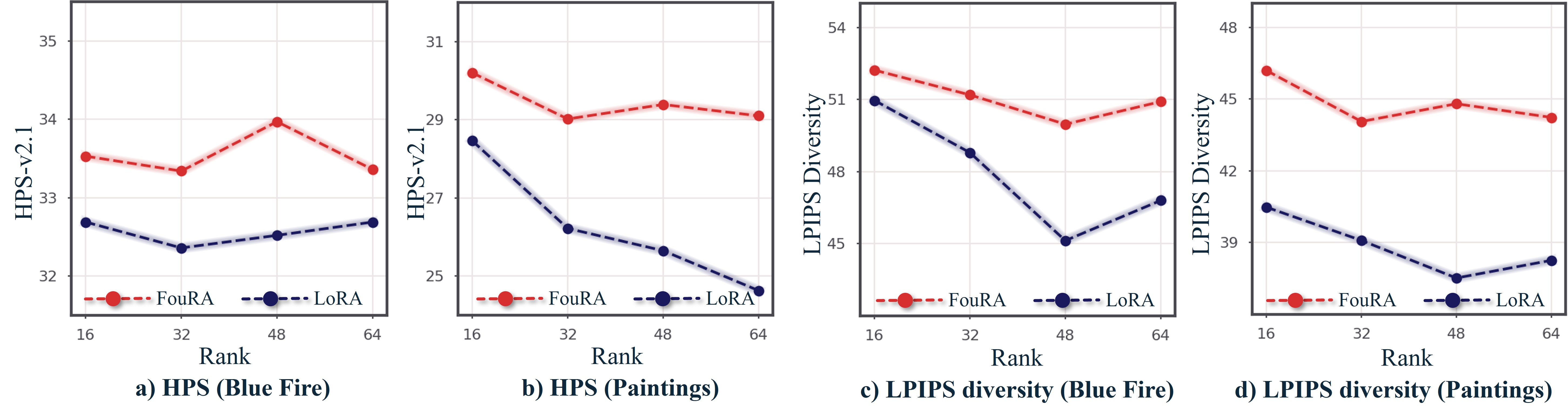}
    \caption{\textbf{HPS-v2.1 scores for each adapter across ranks.} FouRA continues to outperform LoRA as the rank increases for both Paintings and Blue Fire datasets.}
    \label{fig:rank_graph}
    
\end{figure*}

\noindent \textbf{Varying the Rank in Text-to-Image Stylized Generation:}
In Fig.~\ref{fig:rank_graph}, we investigate the impact of FouRA over varying values of input rank, and compare with LoRA. We observe that rank is a highly sensitive parameter for LoRA. However, the HPS scores across ranks for FouRA are higher than the highest HPS score acheived at any rank by LoRA, highlighting the effect of gating in frequency domain. This helps FouRA to avoid underfitting as the rank reduces and overfitting as it increases. Furthermore, FouRA generates a diverse set of images across all ranks.

\section{Conclusion}
\label{sec:conclusion}

\vspace{-0.8 em}
In this paper, we proposed FouRA, a parameter efficient fine-tuning method within the frequency domain. Through extensive experiments and rigorous analysis, we showed that FouRA successfully solves the problems related to data copying and distribution collapse while significantly improving the generated image quality over LoRA. We also present an intensive study on the impact of compact representation of Low rank subspaces in transformed domain. Further, we showed that FouRA can leverage our proposed adaptive mask ranking approach and further push the generalization capabilities of PEFT models without under-fitting. Additionally, we demonstrated the efficacy of FouRA in merging two concepts, as the frequency domain acts as a decorrelated subspace for multiple adapters. Assessing the performance of FouRA, we feel encouraged to think that frequency domain fine-tuning of adapters will potentially be a popular research direction in the coming years.


\bibliographystyle{plain}
\bibliography{main}

\clearpage

\appendix
\appendixpage
\counterwithin{figure}{section}
\counterwithin{table}{section}
\section{Contents}
\label{sec:SuppleIntro}
As part of the supplementary materials for this paper, we share our Implementation details, show extended qualitative and quantitative results and provide additional theoretical analysis for our proposed approach. The supplementary materials contain: 

\begin{easylist}[itemize]
@ Extended Theoretical Analysis
    @@  Proof of Singular Value Decomposition Analysis Lemma \ref{lem:svd}
    @@  Proof of Sparsity Lemma \ref{lem:sparsity} 
    @@  Subspace Analysis 
    @@  Merging of Adapters
    @@  Learning disentangled representations
@ Implementation details and hyperparameters for all experiments
    @@ Datasets 
    @@ Hyperparameters 
@ Interpretations for learnt metrics (HPS-v2.1 and LPIPS diversity)
@ Additional experiments for text-to-image stylization.
        @@ Computational Complexity
        @@ Performance on Unseen Concepts for Text-to-Image Stylization
        @@ Effect of varying the frequency transform
        @@ Comparisons: 2D FFT on the tokens vs 1D FFT on token embeddings
        @@ Plots for quantiative metrics in Text-to-Image Stylization
        @@ Additional Visual Results on Text-to-Image Stylization
@ Additional Experiments for Text-to-Image Editing using Concept Sliders
@ Societal Impacts

\end{easylist}

\section{Theoretical Analysis} \label{appendix_theory}

\subsection{Proof for Lemma~\ref{lem:svd}}
\label{sec:appendix_svd}

In this section, we provide the proof for Lemma~\ref{lem:svd} of the main text.

\textbf{Lemma~\ref{lem:svd}.} \textit{Considering two adapters $\Delta\mathbf{W_1}$ and $\Delta\mathbf{W_2}$ and their corresponding sets of singular values $\{\sigma_{1,i}\}$ and $\{\sigma_{2,i}\}$. The adapter $\Delta\mathbf{W_1}$, will admit $r$ rank approximation with lower error than $\Delta\mathbf{W_2}$ if $\sigma_{1,i} < \sigma_{2,i}$ for all $i \geq r$.
}
\begin{proof}
Let $\mathbf{D_{1,r}}$ and $\mathbf{D_{2,r}}$ be diagonal matrices corresponding a rank $r$ approximation of $\Delta\mathbf{W_1}$ and $\Delta\mathbf{W_2}$ respectively. The reconstruction errors $\mathbf{E_{1,r}}$ and $\mathbf{E_{2,r}}$ for these approximations are computes as follows:

\begin{equation}
    \mathbf{E_{1,r}} = \Delta\mathbf{W_1} - LR_r(\Delta\mathbf{W_1}) = \mathbf{U_1}\mathbf{D_1}\mathbf{V_1^T} - \mathbf{U_1}\mathbf{D_{1,r}}\mathbf{V_1^T} 
\end{equation}
\begin{equation}
    \mathbf{E_{2,r}} = \Delta\mathbf{W_2} - LR_r(\Delta\mathbf{W_2}) = \mathbf{U_2}\mathbf{D_2}\mathbf{V_2^T} - \mathbf{U_2}\mathbf{D_{2,r}}\mathbf{V_2^T}
\end{equation}

A matrix $\Delta\mathbf{W}$ can be written as the sum of it's right and left 1-D singular vectors $\mathbf{u}$ and $\mathbf{v}$ as follows:

\begin{equation}
    \Delta\mathbf{W} = \mathbf{U}\mathbf{D}\mathbf{V^T} = \sum_{i=1}^{min(k_1, k_2)} \sigma_i\mathbf{u}\mathbf{v^T}
\end{equation}

Hence, we rewrite the reconstruction errors $\mathbf{E_{1,r}}$ and $\mathbf{E_{2,r}}$ as a sum of the product of their 1-D singular vectors. 

\begin{equation}
    \mathbf{E_1} = \sum_{i=1}^{min(k_1, k_2)} \sigma_{1,i}\mathbf{u_1}\mathbf{v_1^T} - \sum_{i=1}^{r} \sigma_{1,i}\mathbf{u_1}\mathbf{v_1^T} = \sum_{i=r+1}^{min(k_1, k_2)} \sigma_{1,i}\mathbf{u_1}\mathbf{v_1^T}
\end{equation}

\begin{equation}
   \therefore \mathbf{E_2} = \sum_{i=r+1}^{min(k_1, k_2)} \sigma_{2,i}\mathbf{u_2}\mathbf{v_2^T}
\end{equation}

Following the Eckart-Young theorem~\cite{eckart1936approximation} and theorem 4.95 in Mathematics for Machine Learning~\cite{deisenroth2020mathematics}, the value of the norm of reconstruction error is given as:
\begin{equation}
    \|\mathbf{E_1}\| = \left\Vert\sum_{i=r+1}^{min(k_1, k_2)} \sigma_{1,i}\mathbf{u_1}\mathbf{v_1^T}\right\Vert = \sigma_{1, r+1}
\end{equation}

Hence the difference of reconstruction errors is computed as follows:

\begin{equation}
    \|\mathbf{E_{2,r}}\| - \|\mathbf{E_{1,r}}\| = \sigma_{2, r+1} - \sigma_{1, r+1}
\end{equation}

We know $\sigma_{2, r+1} > \sigma_{1, r+1}$. Hence, we prove that $\|\mathbf{E_{2,r}}\| > \|\mathbf{E_{1,r}}\|$.

\end{proof}

Here it is important to note an adapter with lesser eigenvalue spread there will exist an $r$ rank approximation such it has a lower approximation error than adapter with wider eigenvalue spread. However, the rank $r$ should follow in lemma above. Further, it is important note the low rank adapter with a lower approximation error would estimate the noise closer to optimal estimate and will converge to de-noised image with improved perception scores. 

\subsection{Proof for Lemma~\ref{lem:sparsity}}
\label{sec:appendix_sparsity}

In this section, we provide a proof for Lemma~\ref{lem:sparsity} 
 and Corollary~\ref{lem:f_sparsity} of the main text.

\textbf{Lemma~\ref{lem:sparsity}.} \textit{Consider an adapter $\Delta\mathbf{W}$ with a rank higher than the required rank to fit a training data distribution. The upper-bound of generalization error $\mathcal{R}$ for fine-tuning this adapter reduces as the effective rank of the adapter reduces. After reducing to a certain value of effective rank, the upper-bound of generalization error will increase as rank reduces further.
}

\textbf{Corollary~\ref{lem:f_sparsity}.} \textit{
 Additionally, the generalization bound is more stable when the singular value distribution of adapter weights $\Delta\mathbf{W}$ is more compact.
}

\begin{proof}
    Consider $\mathcal{A}$ as a learning algorithm for finetuning our adaptation weights $\Delta\mathbf{W}$, and $\mathbf{S}$ is our training set of length $n$. Additionally, consider the ratio of effective rank to original rank as $p$ (where $1-p$ is a sparsity parameter). The LoRA Generalization error upper-bound for $\mathcal{A}$ can be computed from Pointwise Hypothesis Stability equations (Theorem 2 of~\cite{fu2023effectiveness}). We have for a constant $C$ with a probability $1 - \delta$,

    \begin{equation}
        \mathcal{R}(\mathcal{A}, S) < \mathcal{\hat{R}}(\mathcal{A}) + \sqrt{\frac{C^2 + \frac{24C\rho^2}{\lambda_{min} + 2(1 - p)}}{2n\delta}}
    \end{equation}

    Here, $\mathcal{\hat{R}}(\mathcal{A}, S)$ represents the emperical error, and $\lambda_{min}$ represents the minimum eign-value of the loss Hermitian matrix. For finetuning tasks, $\lambda_{min}\approx0$ for a loss Hermitian matrix which is well behaved as the model has been trained, as observed by~\cite{sagun2016eigenvalues}.
    
    Based on the observations of~\cite{lin2024lora, fu2023effectiveness}, and the above equation, we can observe that the generalization error reduces as the sparsity increases when the effective rank ratio $p$ is low, and sparsity ($1-p$) is relatively high. 
    
    As effective rank increases and sparsity($1-p$) reduces, if the length of data distribution is small, there is a high risk of overfitting. 
    
    However, as effective rank reduces and sparsity increases, there will come a point when the number of trainable parameters are much lower than what is required for representing the training data distribution, leading to underfitting. Hence, there exists an \textbf{optimal} effective rank, proving Lemma~\ref{lem:sparsity}.   

    The optimal effective rank is driven by the generalization error. For highly sparse representations, the empirical error $\mathcal{\hat{R}}(\mathcal{A}, S)$ dominates over the second term, as it increases significantly.

    From Lemma~\ref{lem:svd}, we know that if the singular value spread of $LR_r(\Delta\mathbf{W})$ contains a more compact representation, the reconstruction error from the $r$-rank subspace is reduced. Hence, the training objective $\mathcal{\hat{R}}(\mathcal{A}, S)$ reduces.
    
    A consequence of this reduction in error signifies that the weights can potentially achieve higher generalization capability by even further sparsification, before $\mathcal{\hat{R}}(\mathcal{A}, S)$ starts dominating the generalization error bound.

    Hence, model weights which can be represented in compact singular value representations can achieve a lower generalization error by further increasing sparsity, proving Corollary~\ref{lem:f_sparsity}. 
    
\end{proof}

\subsection{Subspace analysis}
\label{sec:decorr}

\begin{figure*}[h]
    \centering
    \includegraphics[width=0.5\textwidth]{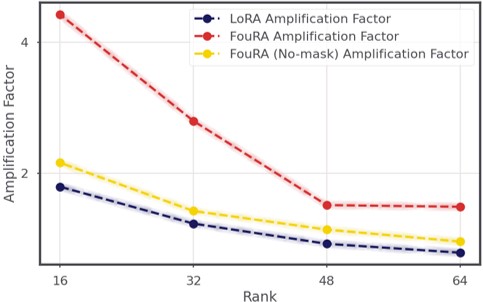}
    \caption{\textbf{Amplification Factor of FouRA v/s LoRA:} As the computed Amplification Factor referred to in~\ref{sec:decorr} is higher in case of FouRA, we justify the learnt representations are more de-correlated from the base weights.}
    \label{fig:amp_factor}

\end{figure*}

In Section~\ref{sec:experiments}, we demonstrate that the fine tuned FouRA adapter performs significantly better than LoRA. In this Section, we attempt to analyze the performance of adapters in terms of the correlation of the subspaces of the base model and that of the adapter. The analysis follows the approach discussed in \cite{lora}. We project the base model weights $\mathbf{W_0}$ onto the $r$-dimensional subspace of our finetuned adapters $\Delta\mathbf{W}$. The projection of base matrix $\mathbf{W_0}$ on to the subspace of the adapter is $ \mathbf{U}^{T}\mathbf{W_0}\mathbf{V}^{T}$, where $\mathbf{U}/\mathbf{V}$ are the left and right top-$r$ singular vectors of $\Delta\mathbf{W}$.  As defined in~\cite{lora},  $\frac{\|\Delta\mathbf{W}\|_{F}}{\| \mathbf{U}^{T}\mathbf{W_0}\mathbf{V}^{T}\|_{F}}$ is the amplification factor, a measure of the subspaces emphasised in the adapter $\Delta\mathbf{W}$ when compared with base weights $\mathbf{W_0}$. Between two adapters of the same rank, a higher amplification factor effectively corresponds to the amount of information learned by the adapter, which is orthogonal to the model weights. In table~\ref{tab:amp_factor}, we analyze the amplification factors of FouRA and LoRA at rank=32. This is an average over all the adaptors of finetuned UNet model. Observe that FouRA Amplifies the learnt subspaces by factor >2x as compared to LoRA. Hence, FouRA weights are more de-correlated from the pretrained base model weights. Additionally, higher emphasis on the set of learnt subsapces enables the learning of new tasks without catastrophic forgetting. Figure~\ref{fig:amp_factor} shows further analysis of learnt subspaces over multiple ranks.

\begin{table*}[h]
    \centering
    \scalebox{1}{
    \fontsize{7.0pt}{5.75pt}\selectfont
    \begin{tabular}{l|ccc} 
    \toprule
        & $\mid\mid\Delta w \mid\mid_{F}$ & $\mid \mid U^{T}WV^{T} \mid \mid_{F}$ ($\downarrow$) & $\frac{\mid\mid\Delta w \mid\mid_{F}} {\mid \mid U^{T}WV^{T} \mid \mid_{F}}$ ($\uparrow$) \\ [0.5ex] 
    \hline
    \midrule
     LoRA & $1.07$ & $0.95$ & $1.2$ \\ 
     \rowcolor{Gray} FouRA & $0.32$ & $0.81$ & $2.8$ \\
    \bottomrule
\end{tabular}
}
    \caption{\textbf{Amplification Factor Analysis.} Average amplification factor components over all layers of the diffusion UNet with Rank=32 LoRA and FouRA.} 
    \label{tab:amp_factor}
\end{table*}

\subsubsection{Merging adapters}

Recent works~\cite{shah2023ziplora} demonstrate joint-adapter training for effectively merging multiple low-rank adapters. In Section~\ref{sec:experiments}, we demonstrate the ability of the FouRA module to merge multiple adaptors in a way which retains both their capabilities with high fidelity.

\begin{prop}
\label{prop:merge}
Considering two adapters $\Delta\mathbf{W_1}$ and $\Delta\mathbf{W_2}$. The linear combination of both these adaptors tends to generate results which retain the capabilities of both the adapters, if the norm of the projection of $\Delta\mathbf{W_1}$ on the subspace of $\Delta\mathbf{W_2}$, computed as $ \|\mathbf{U_2}^{T}\Delta\mathbf{W_1}\mathbf{V_2}^{T}\|$ is lower. Here, $\mathbf{U_2}/\mathbf{V_2}$ are the singular vectors of $\Delta\mathbf{W_2}$.
\end{prop}

We provide analysis in Table~\ref{tab:projection_adapter_merging} complementing Proposition~\ref{prop:merge}, and demonstrating how FouRA has a greater tendency to disentangle two adapters, making it highly effective for multi-adaptor fusion without joint training. We computed the Norm of the projections FouRA adapter weights trained on one subtask, onto the weights trained on another subtask, and compared it to LoRA projection norms. We analyzed the correlation between weights of three tasks: \textit{BlueFire}, \textit{Paintings} and \textit{3D.} As observed from the numbers, FouRA projection norms are much lower, suggesting a higher number of orthogonal subspaces for FouRA projections. This aligns with Table~\ref{tab:merge_styles} and Figure~\ref{fig:adapter_merging} of the main text, where we observe that FouRA is successfully able to retain the capabilities of both adapters after the merge.

\begin{table*}[h]
    \addtolength{\tabcolsep}{-2pt}
    \centering
    \fontsize{7.0pt}{5.75pt}\selectfont
    \begin{tabular} {  cc|cc }
    \toprule
    \textbf{Dataset 1} & \textbf{Dataset 2} & \textbf{LoRA Projection Norm($\downarrow$)} & \textbf{FouRA Projection Norm ($\downarrow$)} \\\hline
    \midrule
    \textit{BlueFire}  & \textit{Paintings} & 0.40 & \textbf{0.25} \\
    \textit{BlueFire} & \textit{3D} & 0.39 & \textbf{0.27} \\
    \textit{3D }& \textit{Paintings }& 0.47 & \textbf{0.32} \\

    \bottomrule
\end{tabular}
    \caption{Norm of projection of adapter weights trained on task 1, over adapter weights trained on task 2, calculated as $ \|\mathbf{U_2}^{T}\Delta\mathbf{W_1}\mathbf{V_2}^{T}\|$. Observe that FouRA has a lower Projection Norm, } 
    \label{tab:projection_adapter_merging}
\end{table*}

\subsection{Learning disentangled representations}
\label{sec:appendix_disentangled}
Given  $z_{in}, z_{out} \in \mathcal{R}^{d \times k_1}$ from \eqref{eq:freq},  and let the input have three attributes that can be represented as   $z_{in} = [z_{race}, z_{age}, z_{gender}]$, the autocorrelation matrix at the output of FouRA layer can be written as 

\begin{equation}
  \begin{aligned}
        \mathbf{R}_{d \times d} = \mathbf{z}_{out} \mathbf{z}_{out}^{T} =\mathbf{z_{in}} ( \mathbf{W_{0}} + \Delta W)( \mathbf{W_{0}} + \Delta W)^{T} \mathbf{z_{in}^{T}}   \\
                   = \mathbf{z_{in}} \mathbf{W_{0}}\mathbf{W_{0}}^{T}\mathbf{z_{in}^{T}} + \textcolor{blue}{\mathbf{z_{in}} \Delta W \Delta W^{T}\mathbf{z_{in}^{T}}} + \mathcal{F} ( \mathbf{W_{0}}\Delta W^{T}, \mathbf{z_{in}} )
    \end{aligned}  
\end{equation}

From \ref{fig:amp_factor}, we established that the overlap of subspaces between low rank in transform domain $\Delta W$ and base matrix $\mathbf{W}$ is smaller at lower rank. In addition, in frequency domain, the term in the middle (in \textcolor{blue}{blue}) computes the autocorrelation between the subspaces. From \cite{beaufays1993simple}, this term is almost diagonal making the dot product  $ <z_{out}^{race}, z_{out}^{gender}> \approx 0$ or $<z_{out}^{race}, z_{out}^{age}> \approx 0$. Thus the weights for each attribute is poised to be learned independently. To verify this, In the experiments section, we motivate the idea of using foura to edit concepts while preserving the attributes of an image using concept sliders \cite{gandikota2023concept}

\section{Implementation Details} \label{appendix_impl}

\subsection{Datasets} \label{language-dataset}
\subsubsection{LLM Benchmark: GLEU}
We have performed the LLM study on six of the GLUE benchmarks - CoLA, SST-2, MRPC, STS-B,
MNLI, and QNLI. GLEU benchamrk has been widely used for natural language understanding. All the dataset and task described in the Table \ref{tab: gleu} is being utilized from Huggingface Datasets and each task has its own respective evaluation metric. We have described the train and test split of each of the task along with the respective evaluation metric in Table \ref{tab: gleu}.

\begin{table*}[h]
    \addtolength{\tabcolsep}{-2pt}
    \centering
    \fontsize{7.0pt}{5.75pt}\selectfont
    \begin{tabular} {  lccc }
    \toprule
    \textbf{Dataset} & \textbf{\#Train} & \textbf{\#Val} & \textbf{Metric} \\\hline
    \midrule
    CoLA & 8.5K & 1043 & Mcc \\
    \midrule
    SST-2 & 67K & 872 &  Acc \\
    \midrule
    MRPC & 3.7K & 408 &  Acc \\
    \midrule
    STS-B & 5.7K & 1.5K &  Corr\\
    \midrule
    MNLI & 393K & 9.8K &  Acc(m/mm)\\
    \midrule
    QNLI & 105K & 5.5K & Acc \\
    \bottomrule
\end{tabular}
    \caption{GLUE Benchmark} 
    \label{tab: gleu}
\end{table*}

\subsubsection{Style Transfer Datasets}\label{style-transfer-details}
In this section, we provide more details on the four style transfer datasets we use for vision adaptation experiments. We followed the licensing terms for every dataset which was curated.



\textbf{BlueFire (Training):}
The \textit{BlueFire } dataset is created by collecting images from open public domain and consist of 6 concepts - car, dragon, bird, fox, man and castle. The dataset has a total of 
54 images covering all the concepts. 

\textbf{BlueFire (Validation):}
The \textit{Bluefire} validation set consists of 30 curated text prompts, of which 9 prompts contain one of 6 categories on which the model was trained, and the remaining 21 prompts correspond to categories which the low-rank adapter has not been fine-tuned on. These contain categories such as: (football, monster, sword, chess rook, lion, tiger, dog, cat, koala, panda).

For all training experiments validating on this dataset, we produce 30 images per prompt, varying the input seed. Hence, the HPS analysis is over 900 image and LPIPS-diversity analysis is over ~14500 image pairs.

\textbf{Paintings:}
On similar lines, the \textit{Paintings }dataset is also a collection of images from public domain (CC0 license). The dataset has a total of 90 images cover 9 concepts - fire, bird, elephants, ship, horse, flower, woman, man and tiger. 

\textbf{Paintings (Validation):}
The \textit{Paintings} validation set consists of 21 curated text prompts, of which 9 prompts contain one of 9 categories on which the model was trained, and the remaining 12 prompts correspond to categories which the low-rank adapter has not been fine-tuned on. These contain categories such as: (lion, tiger, dog, cat, koala, panda, and other landscapes)

\textbf{Paintings merged with BlueFire (Validation):}
The evauation set for merging \textit{Paintings} and \textit{Bluefire} consists of 18 curated text prompts. These contain categories such as: (fox, bird, lion, tiger, dog, cat, koala, panda, and other landscapes)

For all training experiments validating on this dataset, we produce 30 images per prompt, varying the input seed. Hence, the HPS analysis is over 440 image and LPIPS-diversity analysis is over ~8750 image pairs.

\textbf{Origami:}
The \textit{Origami }dataset is also a collection of origami images from public domains. The dataset has a total of 52 images covering 7 concepts - bird, boat, flower, cat, dog, fox and house.

\textbf{3D:}
The \textit{3D } dataset is also a collection of images from public domains. These images are animated images showing 3D concepts. The dataset has a total of 30 images covering 6 concepts - boy, girl, astronaut, cat, dog, elephant, dog and building.

\paragraph{Concept Sliders:}
For concept sliders, we train and evaluate on three different concepts as shown in Table~\ref{tab:concept_data}. The evaluation set for each concept consists of 400 examples, over 10 seeds, essentially validating over 4000 images per concept. We follow the method in \cite{sliders}

\vspace{1.2 em}
    
\begin{table*}[h]
    \addtolength{\tabcolsep}{-2pt}
    \centering
    \fontsize{7.0pt}{5.75pt}\selectfont

\begin{tabular}{ l|cccc } 
\toprule
 \textbf{Concept} & \textbf{Positive prompt} & \textbf{Negative prompt} & \textbf{\# Training Attributes} & \textbf{\# Val. Attributes} \\
    \hline
    \midrule
    \textit{Age} & very old, wrinkly, gray hair, aged skin & very young, smooth skin, youthful & 20 & 400 \\ 
    \textit{Surprise} & looking surprised, wide eyes, open mouth &  looking calm, neutral expression & 20 & 400 \\ 
    \textit{Hair}& curly hair, wavy hair  & straight hair & 20 & 400 \\ 
    \hline
\end{tabular}
    \caption{Dataset statistics for Concept Slider Experiments} 
    \label{tab:concept_data}

\end{table*}

\subsection{Hyper-parameters and Implementation details for all experiments}

\textbf{Text-to-image style transfer}

We used the kohya-ss\footnote{https://github.com/kohya-ss/sd-scripts} repository for finetuning models for the text-to-image stylization task. For the masking we follow the approach for soft gating in \footnote{https://github.com/prachigarg23/Memorisation-and-Generalisation-in-Deep-CNNs-Using-Soft-Gating-Mechanisms}. For each task, we trained both LoRA and FouRA adapters with the same set of hyperparameters. We trained using 4 NVIDIA A100 GPUs, for 100 epochs at at batch size of 8. Our initial learning rate was set to $1e^{-4}$ for UNet and $5e^{-5}$ for the text encoder. LoRA and FouRA modules are applied in the default places for stable-diffusion-v1.5 backbone, same as in HuggingFace Diffusers. We trained using two sets of weights, the base sd-1.5\footnote{https://huggingface.co/runwayml/stable-diffusion-v1-5} from runwayML, and RealisticVision3.0\footnote{https://huggingface.co/spaces/Thafx/sdrv30}. For some ablation studies, we varied the rank between {16, 32, 48, 64}. In all the remaining experiments, we set the rank at 64 unless stated otherwise. Additionally, we set the Realistic Vision weights as our default for all experiments.

For quantitative evaluation, we observed the HPS-v2.1 and LPIPS-Diversity metrics at a range of values between $[0, 1]$ for adapter strength $\alpha$. In all quantitative evaluations, we averaged over the same set of 30 seeds $\{0, 1, 2, .... 29\}$.

\textbf{Image editing using Concept Sliders}
\paragraph{Single slider:}  The training data used  in these experiments were curated from  \cite{gandikota2023concept} . We used the repository  \footnote{https://github.com/rohitgandikota/sliders} for finetuning the adapters. We train across 20 different attributes spanning different genders and races and other person attributes for each concept.  The learning rate and other hyperparameters are re-used from the repository. 
For all the experiments we fix a rank of 8 and with 50 denoising steps. 
For evaluations, we tested across 400 different examples for 10 seeds on each prompt including unseen categories such as 'doctor' , 'barista', 'cowboy' .  For qualitative analysis, we compare across  strengths $ \in [-6, 6])$.  We also evaluated the inference across different  3 different edit times [750, 800, 850]. 
\paragraph{Composite slider:}  For compositing we use similar setup as in the single slider.  We  compose the score functions using additive guidance. Specifically we weight each score function based on the relative strengths of the adapter during inference.  

\textbf{GLUE benchmark experiments}
We trained the LoRA and SoRA~\cite{ding2023sparse} baselines on the GLUE benchmark using the code and default set of hyper-parameters provided by the authors\footnote{https://github.com/TsinghuaC3I/SoRA}. For training FouRA, we used the same set of hyper-parameters as the LoRA baseline. These are provided in \hyperlink{https://github.com/TsinghuaC3I/SoRA/issues/4}{this issue} in their repository. For all the experiments, we trained using 1 NVIDIA A100 GPU.

For each task, and each baseline, we evaluated on all the samples of the validation set, the size of which is mentioned in Appendix~\ref{tab: gleu}. This is slightly different from the evaluation in~\cite{ding2023sparse}, as the authors originally ran inference only on a subset of the validation set, indicated \hyperlink{https://github.com/TsinghuaC3I/SoRA/issues/7}{here}. Additionally, we used the set of three seeds $\{100, 81, 20\}$, chosen at random, to run all experiments.

\section{Interpretations for Metrics}
\label{sec:metricinterpret}

In the main text, we used two metrics to validate style transfer on text-to-image diffusion models. Both are learnt metrics, i.e. HPS-v2.1~\cite{wu2023human} and LPIPS-Diversity~\cite{zhang2018perceptual}. In this section, we provide reference ranges for both metrics, and how they can be interpreted.

\subsection{LPIPS Diversity}
\begin{figure*}[h]
    \centering
    \includegraphics[width=1.0\linewidth]{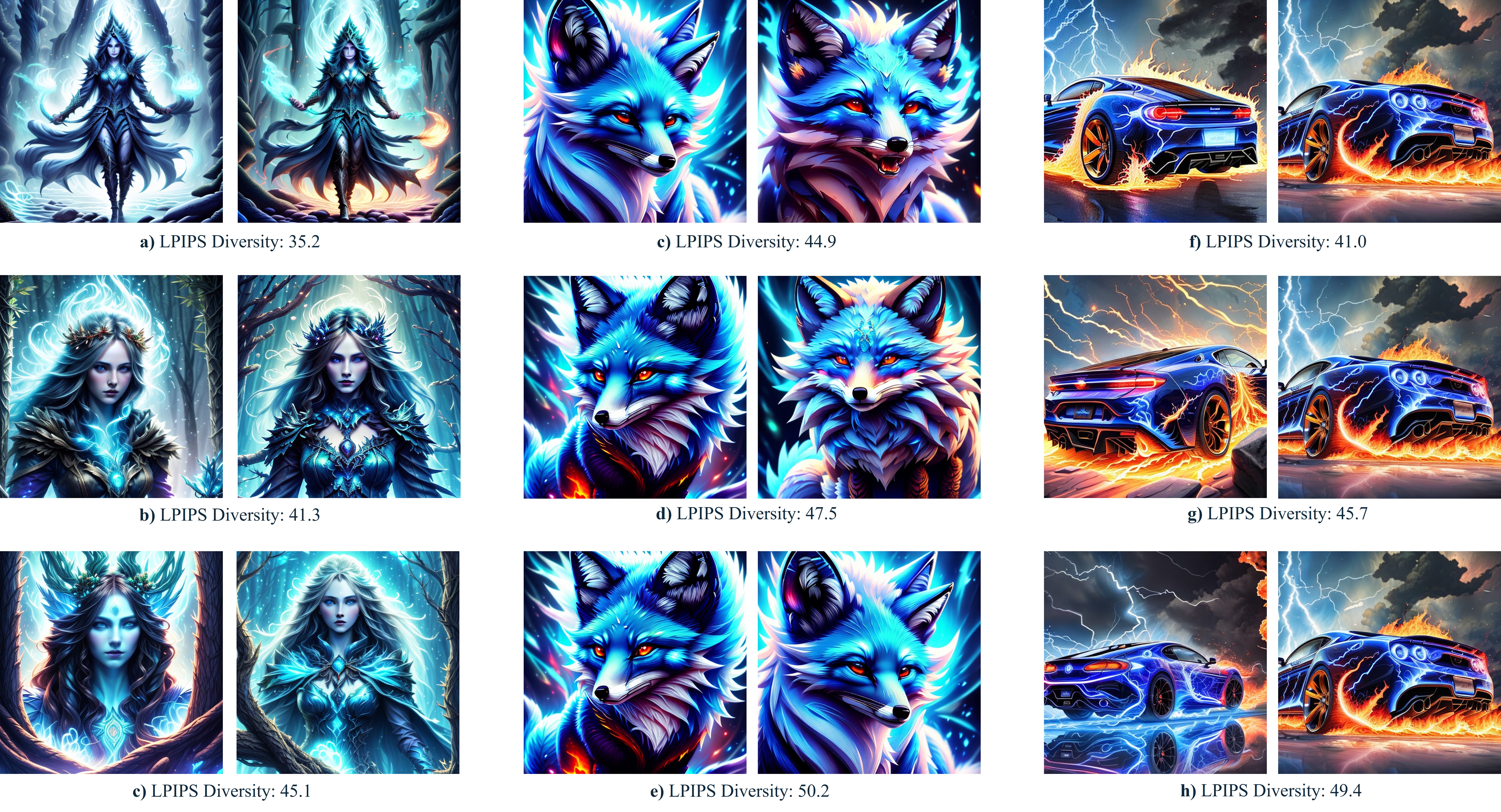}
    \caption{\textbf{Interpretation of the LPIPS Diversity metric}. This figure illustrates the interpretation of LPIPS Diversity, which we used to detect mode collapse. Images which look similar (i.e. sharing the same pose or similar characteristics) tend to generate a lower LPIPS distance.}
    \label{fig:interpret_lpips}
    \vspace{1 em}
\end{figure*}

We compute the LPIPS diversity $\delta_{lpips}$ of a dataset of $n$ images as the average of the LPIPS pairwise distance between $\Comb{n}{2}$ image pairs. In Figure~\ref{fig:interpret_lpips}, we provide reference ranges for LPIPS distance between pairs of images. Notice the images in~\ref{fig:interpret_lpips}a. are very similer. Hence, they generate a low LPIPS score (0.35).  Hence in Table~\ref{tab: t2i_table}, we observe for high values of $\alpha$, as the average LPIPS scores reflect that LoRA produces close to identical images in many case, but FouRA successfully gets rid of this data copying problem. Figures~\ref{fig:interpret_lpips}b. and c. are lesser correlated from each other and hence produce a higher distance. Figures~\ref{fig:interpret_lpips}d.-f. and g.-i. similarly vary from one another with in ascending order of LPIPS diversity scores, which is reflected in the image (The pose of the fox and variations in the fire in car images). The scores in Table~\ref{tab: t2i_table} reflect a gain of 2-6 points in LPIPS diversity between LoRA and FouRA. These are significant improvements in the diversity of generated samples as observed from Figure~\ref{fig:interpret_lpips}.

\subsection{Human Preference Scores}
\begin{figure*}[h]
    \centering
    \includegraphics[width=1.0\linewidth]{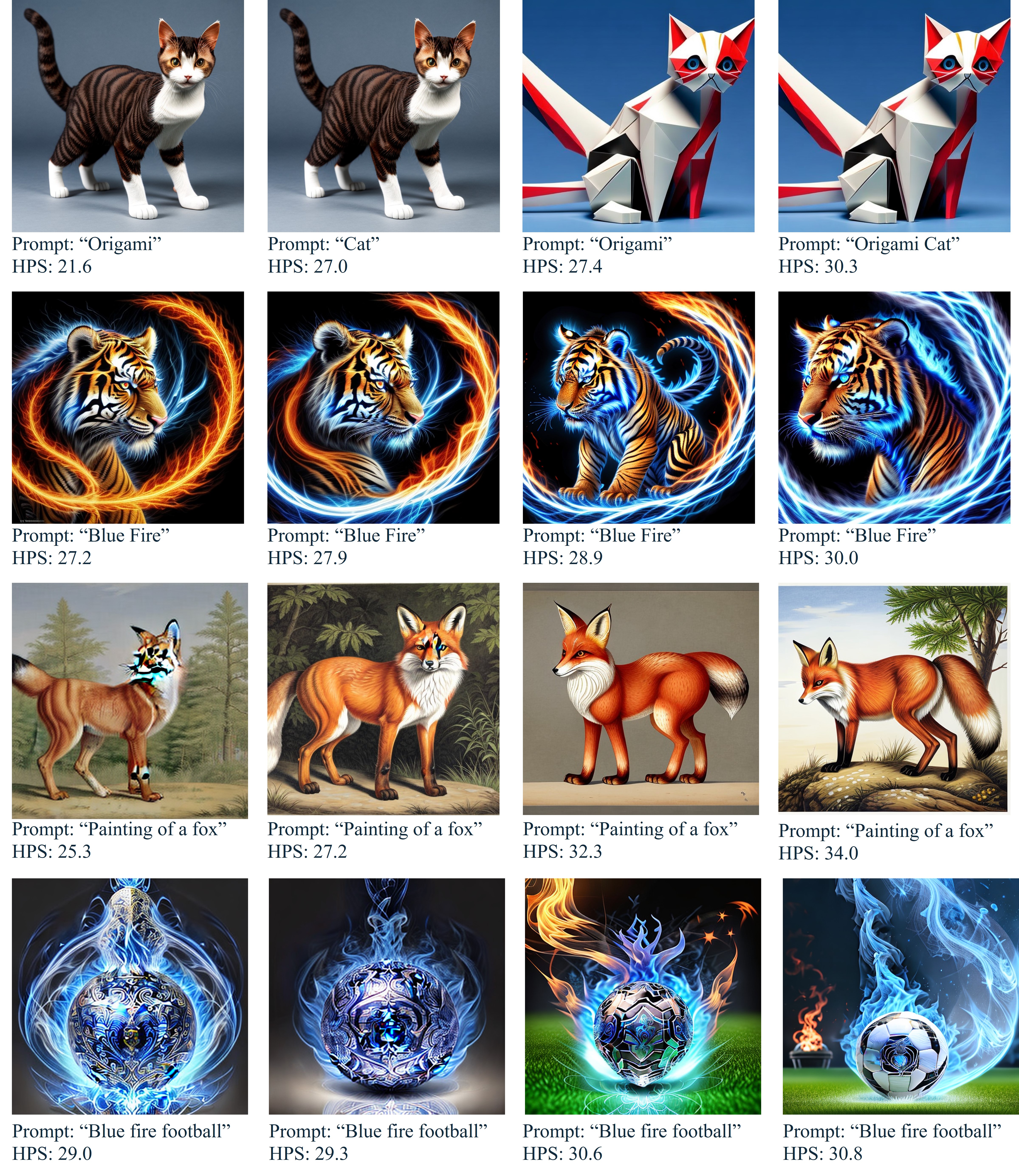}
    \caption{\textbf{Interpretation of the HPS-v2.1 metric}. This figure illustrates the interpretation of HPS scores, which we used to track three key aspects of generated images: 1.Alignment with the prompt, 2.Alignment with the adapter style and 3.Aesthetic quality. Observe that the HPS-v2.1 metric is able to effectively quantify these key aspects of generated images. \textbf{The "Prompt" in this figure corresponds to the input prompt to HPS model for text and image alignment, and may or may not be the prompt used to generate the image}}
    \label{fig:hps_interpret}
    \vspace{1 em}
\end{figure*}

For computing Human Preference Score, we utilized to the v2.1 HPS model provided by the authors~\cite{wu2023human}. Please refer to Figure~\ref{fig:hps_interpret} for reference HPS-v2.1 values. \textbf{Please note that in the Figure~\ref{fig:hps_interpret} the "prompt" corresponds to the input prompt to HPS model, and may or may not be the prompt used to generate the image.}

We used HPS as a metric to track a combination of three key aspects of generated images. 
\textbf{Alignment with the Prompt:} Observe the first row in Figure~\ref{fig:hps_interpret}. For the wrong prompt (e.g. "Origami" for a cat image), the model produces a low HPS score (21.6). However, this score increases as the prompt and image alignment improves.

\textbf{Strength of the adapter:} Observe the second row in Figure~\ref{fig:hps_interpret}. The prompt we fed into HPS is the name of the adapter(\textit{blue fire}). Notice how the HPS values increase for increase in the adapter strength.

\textbf{Image Quality:} Observe the third row in Figure~\ref{fig:hps_interpret}. HPS scores can successfully differentiate between images with high and low aesthetic quality.

Thus the, HPS provides us with a quantifiable metric for all the three aspects over we wish to evaluate our finetuned adapters. Moreover, the fourth row in Figure~\ref{fig:hps_interpret} shows how the HPS can effectively track all these three aspects at once. Hence, the prompt we feed to the HPS model to evaluate an image is a combination of the name of the adapter and the prompt used for generating the image. E.g. the prompt used to evaluate image generated by "dog in space" with the adapter \textit{BlueFire}, is "blue fire dog in space."

This method also works well for evaluating the merging of two adapters. We simply add both the adapter names in the prompts while evaluating their HPS scores.

\section{Additional Experiments on Text-to-Image stylization}
\label{sec:app_t2i}
\subsection{Computational Analysis}
\label{computational_comp}

Table~\ref{tab:comp} provides the computational analysis for FouRA, as compared to LoRA. We provide the \#parameters during inference along with the training time for FouRA. Along with this, we show the HPS-v2.1 scores on the \textit{Blue Fire} validation set. Additionally, we provide the results for a FouRA variant with a fixed gating strategy during inference. FouRA layers with inference-adaptive masking produce an overhead of 0.02\% more than LoRA, as compared to base model weights. However, FouRA with frozen masking can essentially reduce the computational overhead by a factor of 2, and still retain a higher performance than the base LoRA. We used a batch size = 1 on Nvidia A100 GPU for reporting these measurements.
\\
\begin{table*}[h]
    \addtolength{\tabcolsep}{-2pt}
    \centering
    \fontsize{7.0pt}{5.75pt}\selectfont
    \begin{tabular} {  l|cccc }
    \toprule
    \textbf{Adapter} & \textbf{Training Time} & \textbf{\#Parameters (Inference)} & \textbf{$\%$ overhead ($\downarrow$)} & \textbf{HPSv2 score (\textit{BlueFire}) ($\uparrow$)} \\\hline
    \midrule
    None &  & 859.5M & & \\
    \midrule
    LoRA & 1.87 iter/sec & 886.7M & 3.16\% & 32.7 \\
    \rowcolor{Gray} FouRA (Inference-Adaptive Mask) & 2.09 iter/sec & 886.9M & 3.18\% & \textbf{33.4} \\
    \rowcolor{Gray} FouRA (Frozen Mask) & 2.07 iter/sec & 873.1M & \textbf{1.62\%} & 33.1 \\

    \bottomrule
\end{tabular}
    \caption{Computaional Analysis of FouRA. The adapters are all rank=64, and HPS-v2 is computed at $\alpha=1$.} 
    \label{tab:comp}
\end{table*}

\subsection{Additional Ablation Studies}
\subsubsection{Performance on Unseen Concepts for Text-to-Image Stylization}
\label{sec:unseen}
Section~\ref{style-transfer-details} details the distribution of both our validation sets, \textit{Bluefire} and \textit{Paintings}. We split the validation set in seen and unseen concepts during training of the adapter. \textit{Bluefire} contains 21 unseen categories (630 generated images), and \textit{Paintings} contains 12 unseen categories (360 generated images). From Table~\ref{tab:unseen_classes}, we can observe that FouRA has a better generalization capability on unseen classes, as compared to LoRA. This result supplements our Proof for Corollary~\ref{lem:f_sparsity}, essentially confirming that FouRA is able to reduce the upper bound of generalization error.

\begin{table*}[h]
    \addtolength{\tabcolsep}{-1pt}
    \centering
    \fontsize{7.0pt}{5.75pt}\selectfont
    \begin{tabular} {  lc|ccc }
    \toprule
    \multirow{3}{*}{\textbf{Adapter}} & \multirow{3}{*}{\textbf{Dataset}} &  \multicolumn{3}{c}{\textbf{HPSv2 score($\uparrow$)}} \\
        & & $\alpha=1.0$ & $\alpha=0.8$ & $\alpha=0.6$ \\
    \hline
    \midrule
    LoRA  & \textit{Paintings (Unseen)} &  $24.1$ & $27.0$ & $29.7$ \\
    \rowcolor{Gray} FouRA & \textit{Paintings (Unseen)} & $\mathbf{28.5}$ & $\mathbf{30.4}$ & $\mathbf{31.7}$ \\
    LoRA  & \textit{Bluefire (Unseen)} &  $32.5$ & $33.6$ & $33.8$ \\
    \rowcolor{Gray} FouRA & \textit{Bluefire (Unseen)} & $\mathbf{33.2}$ & $\mathbf{34.4}$ & $\mathbf{34.4}$ \\

    \bottomrule
\end{tabular}
    \caption{\textbf{Performance on unseen classes}. Shows that on unseen classes FouRA generalizes better on unseen categories.} 
    \label{tab:unseen_classes}
\end{table*}

\subsubsection{Effect of varying the frequency transform}
Finally, we evaluate the effect of changing the frequency transform between DFT and DCT for our proposed FouRA (see Table~\ref{tab:dft_dct}). First, we observe that both DFT- and DCT-based FouRA models significantly outperform LoRA. Also, both DFT and DCT achieve comparable scores in terms of HPSv2 which means our approach is robust to the type of frequency transforms being used. 
\begin{table*}[h]
    \addtolength{\tabcolsep}{-1pt}
    \centering
    \fontsize{7.0pt}{5.75pt}\selectfont
    \begin{tabular} {  l|ccc|ccc }
    \toprule
    \multirow{3}{*}{\textbf{Transform}} & \multicolumn{3}{c|}{\textbf{LPIPS Diversity($\uparrow$)}} & \multicolumn{3}{c}{\textbf{HPSv2 score($\uparrow$)}} \\
        & $\alpha=1.0$ & $\alpha=0.8$ & $\alpha=0.6$ & $\alpha=1.0$ & $\alpha=0.8$ & $\alpha=0.6$ \\
    \hline
    \midrule
    LoRA  & $38.3$  & $37.8$ & $39.1$ &  $24.6$ & $27.7$ & $30.3$ \\
    \rowcolor{Gray} FouRA DFT  & $44.2$ & $44.7$ & $44.8$ & $\mathbf{29.1}$ & $\mathbf{30.9}$ & $\mathbf{32.2}$ \\
    \rowcolor{Gray} FouRA DCT  & $\mathbf{46.7}$ & $\mathbf{45.5}$ & $\mathbf{45.0}$ & $28.9$ & $30.6$ & $31.9$  \\
    \bottomrule
\end{tabular}
    \caption{\textbf{Effect of varying the frequency transform in FouRA}} 
    \label{tab:dft_dct}
\end{table*}

\subsubsection{Comparisons: 2D FFT on the tokens vs 1D FFT on token embeddings}
\label{sec:appendix_variants}

\begin{figure*}[h]
    \centering
    \includegraphics[width=0.4\textwidth]{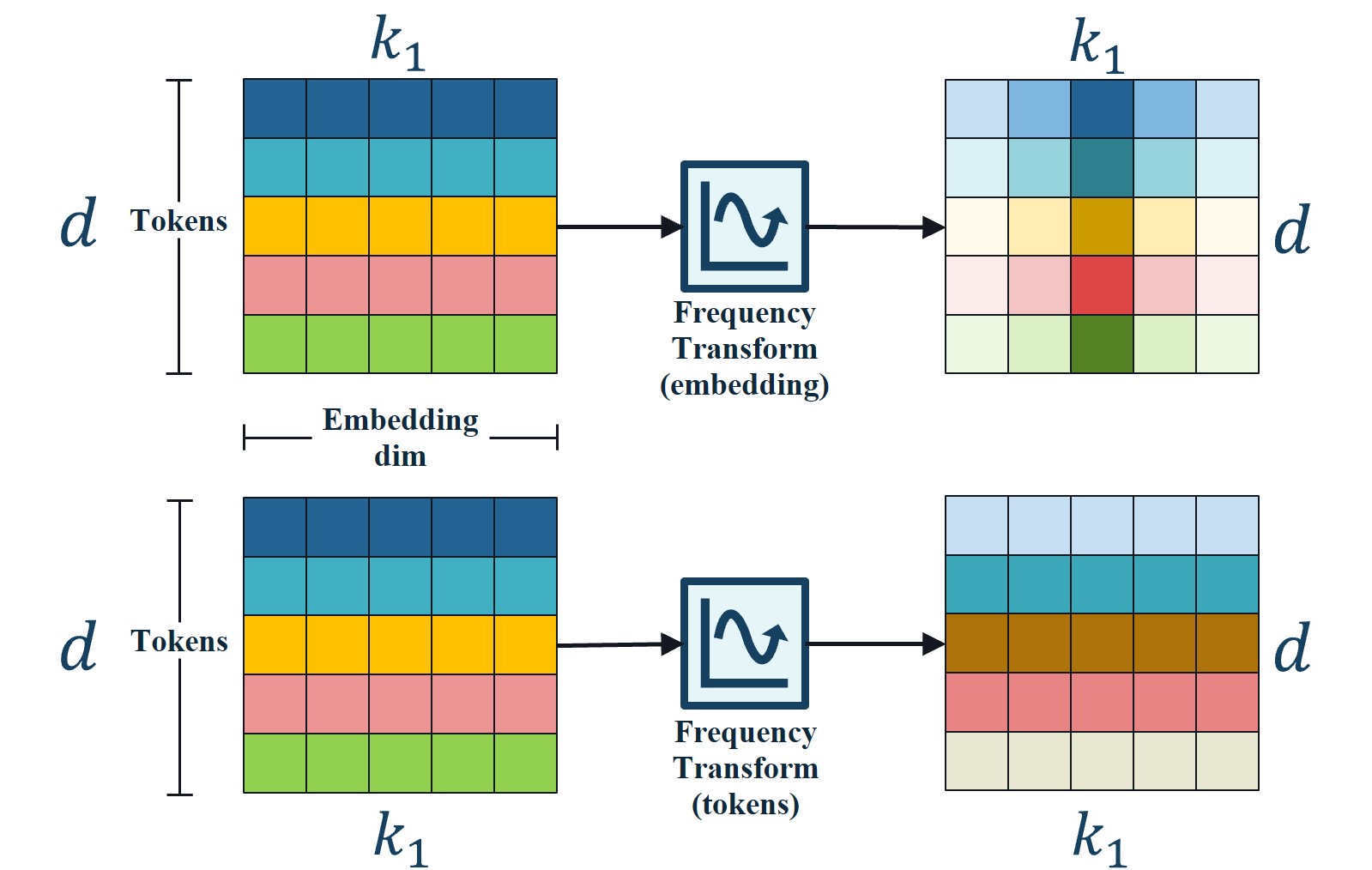}
    \caption{\textbf{Two directions of the proposed Frequency Transform}. 
    FouRA$_{emb}$ computes the frequency transform along the embedding dimension (top), whereas FouRA$_{token}$ computes the frequency transform across all the tokens (bottom).}
    \label{fig:variants}
\end{figure*}

As illustrated in Fig.~\ref{fig:variants}, we proposed two variants of our approach: (1) FouRA$_{emb}$ that computes the frequency transform across the embedding dimension, and (2) FouRA$_{token}$ that computes the frequency transform along the token dimension.

\vspace{1 em}

Table~\ref{tab:t2i_table_variants}, we compare FFT applied on token embeddings with LoRA. We hypothesize that transform done this way might capture variations in local patches of the image. Further as LoRA on vision adaptors generally apply rank reduction in the embedding dimension, applying the same in fourier dimension translates to spectral filtering in the embedding space. For the sake of completeness, we also run experiments to apply transform in the 2D token space, we call this $FouRA_{token}$. In Table~\ref{tab:t2i_table_variants}, we empirically observe that FouRA$_{emb}$ performs better than FouRA$_{token}$. Hence, unless stated otherwise, we set FouRA$_{emb}$ as the default variant of FouRA for our experiments.

\begin{table*}[h]
    \addtolength{\tabcolsep}{-1.5pt}
    \centering
    \fontsize{6.0pt}{4.5pt}\selectfont
    \begin{tabular} {  lc|c|ccc|ccc }
    \toprule
    
    \multirow{2}{*}{\textbf{Style}} & \multirow{2}{*}{\textbf{Base Model}} & \multirow{2}{*}{\textbf{Adapter}} & \multicolumn{3}{c|}{\textbf{LPIPS Diversity($\uparrow$)}} & \multicolumn{3}{c}{\textbf{HPSv2 score($\uparrow$)}} \\

    & & & $\alpha=1$ & $\alpha=0.8$ & $\alpha=0.6$ & $\alpha=1$ & $\alpha=0.8$ & $\alpha=0.6$ \\
    
    \hline
    \midrule
    
    \multirow{3}{*}{Painting} & \multirow{3}{*}{RealisticVision} & LoRA & $38.3 \pm 3.5$ & $37.8 \pm 3.6$ & $39.2 \pm 3.7$ & $24.6 \pm 1.8$ & $27.7 \pm 1.8$ & $30.3 \pm 1.7$ \\
    & & \cellcolor{Gray} FouRA$_{token}$ & \cellcolor{Gray} $\mathbf{44.2 \pm 3.7}$ & \cellcolor{Gray} $44.5 \pm 4.0$ & \cellcolor{Gray} $44.6 \pm 3.9$ & \cellcolor{Gray} $28.4 \pm 1.8$ & \cellcolor{Gray} $30.6 \pm 1.5$ & \cellcolor{Gray} $32.0 \pm 1.4$  \\
    & & \cellcolor{Gray} FouRA$_{emb}$ & \cellcolor{Gray} $\bm{44.2 \pm 3.8}$ & \cellcolor{Gray} $\bm{44.7 \pm 3.9}$ & \cellcolor{Gray} $\bm{44.8 \pm 3.9}$ & \cellcolor{Gray} $\bm{29.1 \pm 1.9}$ & \cellcolor{Gray} $\bm{30.9 \pm 1.6}$ & \cellcolor{Gray} $\bm{32.2 \pm 1.5}$  \\
    \midrule
    \multirow{3}{*}{Blue Fire} & \multirow{3}{*}{RealisticVision} & LoRA & $46.8 \pm 4.0$ & $48.5 \pm 4.0$ & $49.8 \pm 4.2$ & $32.7 \pm 1.6$ & $33.8 \pm 1.4$ & $34.0 \pm 1.5$ \\
    & & \cellcolor{Gray} FouRA$_{token}$ & \cellcolor{Gray} $50.4 \pm 3.0$ & \cellcolor{Gray} $51.6 \pm 3.3$ & \cellcolor{Gray} $52.2 \pm 3.5$ & \cellcolor{Gray} $\bm{33.6 \pm 1.5}$ & \cellcolor{Gray} $34.1 \pm 1.2$ & \cellcolor{Gray} $34.0 \pm 1.4$ \\
    & & \cellcolor{Gray} FouRA$_{emb}$ & \cellcolor{Gray} $\bm{50.9 \pm 3.1}$ & \cellcolor{Gray} $\bm{52.3 \pm 3.2}$ & \cellcolor{Gray} $\bm{53.3 \pm 3.8}$ & \cellcolor{Gray} $33.4 \pm 1.7$ & \cellcolor{Gray} $\bm{34.6 \pm 1.3}$ & \cellcolor{Gray} $\bm{34.5 \pm 1.2}$ \\
    
    \bottomrule
    \end{tabular}
    \caption{FouRA$_{emb}$ vs FouRA$_{token}$ vs LoRA} 
    \label{tab:t2i_table_variants}
\end{table*}

\subsection{Plots for quantiative metrics in Text-to-Image Stylization}
\begin{figure*}[h]
    \centering
    \includegraphics[width=1.0\linewidth]{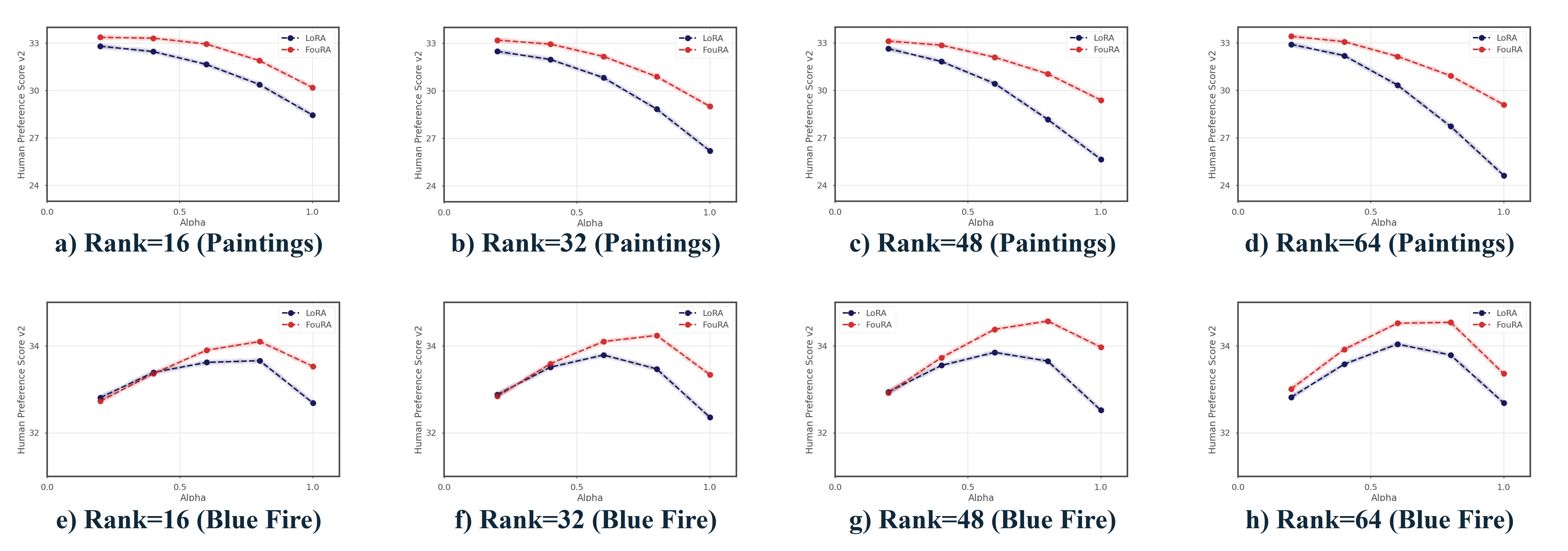}
    \caption{\textbf{Quantitative Evaluations for LoRA v/s FouRA on text-to-image stylization}. We provide plots at ranks $\{16, 32, 48, 64\}$ and adapter strengths $\alpha=\{0.2, 0.4, 0.6, 0.8, 1.0\}$}
    \label{fig:allrank_plots}
    \vspace{- 1 em}
\end{figure*}

In Fig.~\ref{fig:allrank_plots}, we provide HPS and LPIPS-diversity scores at ranks $\{16, 32, 48, 64\}$ and adapter strengths $\alpha=\{0.2, 0.4, 0.6, 0.8, 1.0\}$ for LoRA and FouRA. These plots are using the base weights of Realistic Vision-3.0. These scores are an extenstion to Table~\ref{tab: t2i_table} of the main text. Observe FouRA outperforms LoRA on both metrics, at all ranks.

\subsection{Additional Visual Results on Text-to-Image Stylization}
\begin{figure*}[h]
    \centering
    \includegraphics[width=1.0\linewidth]{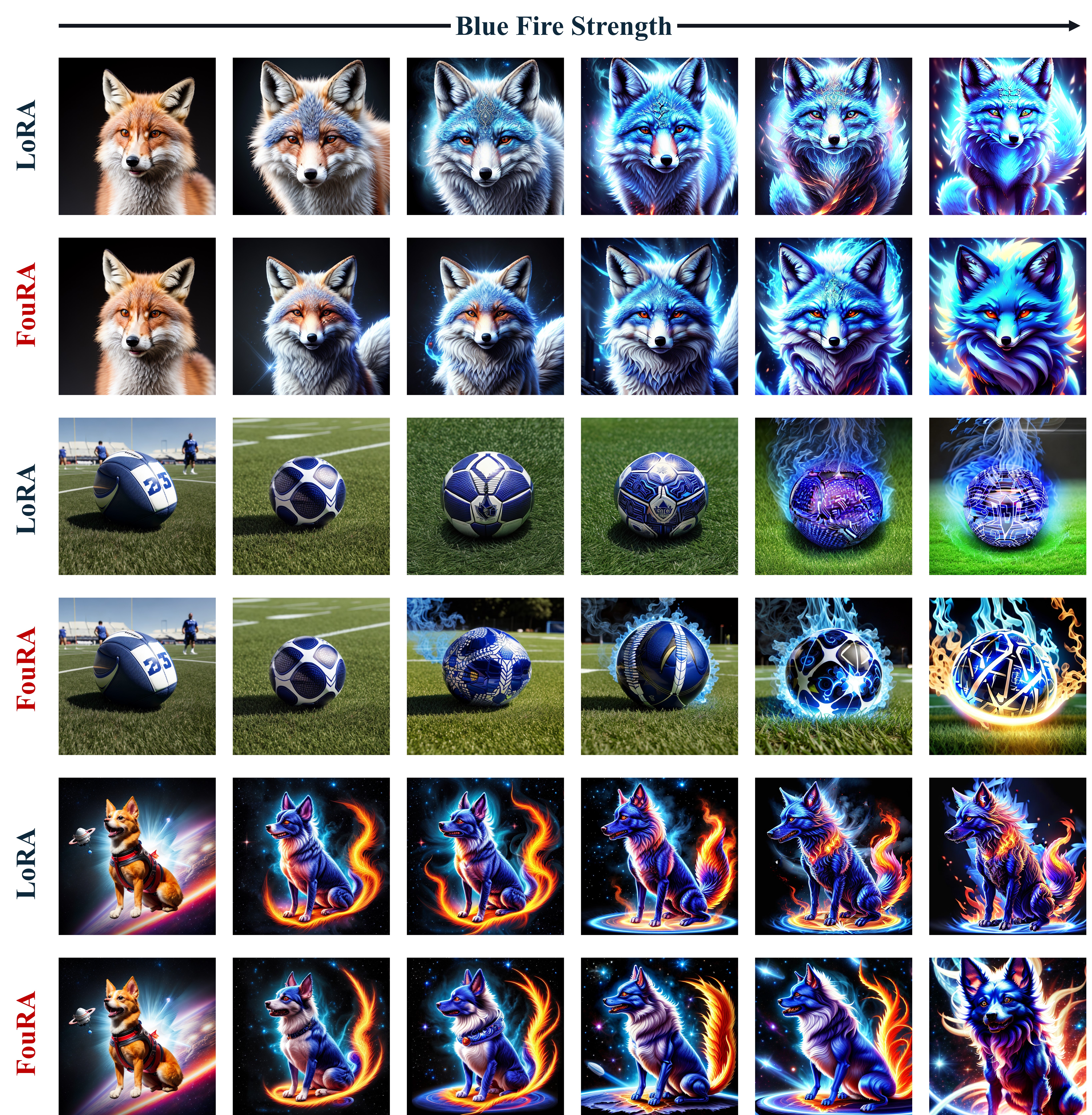}
    \caption{\textbf{Visual Results using \textit{BlueFire}} adapters comparing LoRA and FouRA at varying values of $\alpha$.}
    \label{fig:bluefire_alpha}
\end{figure*}

In Figure~\ref{fig:bluefire_alpha}, we provide additional visual results for FouRA and LoRA finetuning on the \textit{Bluefire} dataset at varying adapter strengths. Within the generated images, the concepts 'Football' and 'Dog' are unseen. As observed, FouRA produces aesthetically appealing images as compared to LoRA in all cases. This is more evident in the 'Football' example. As observed, FouRA can generalize better to new concepts, as compared to LoRA.

\begin{figure*}[h]
    \centering
    \includegraphics[width=1.0\linewidth]{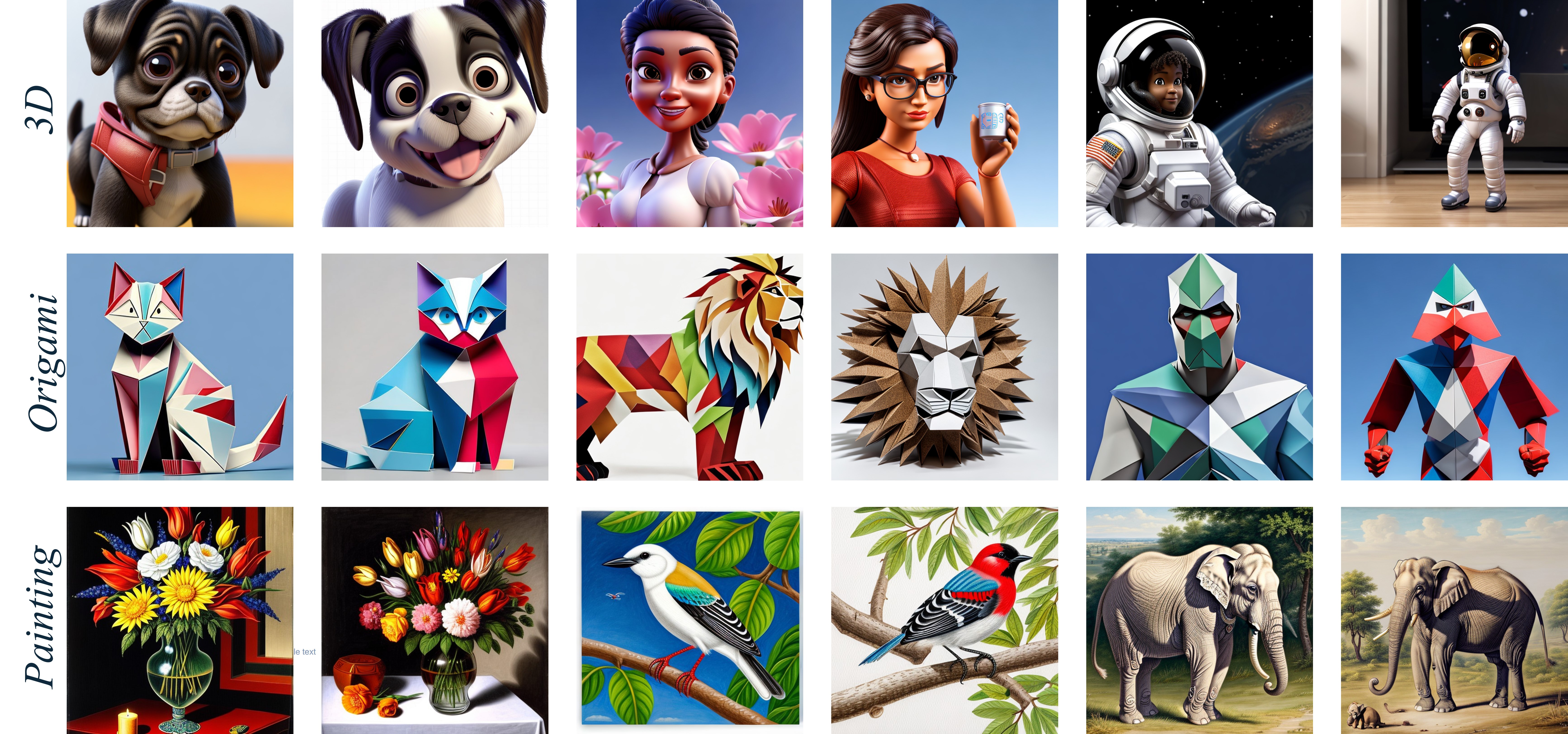}
    \caption{\textbf{Images generated by FouRA} trained on \textit{3D}, \textit{Paintings} and \textit{Origami} datasets.}
    \label{fig:3d_ptg_origami}

\end{figure*}

In Figure~\ref{fig:3d_ptg_origami}, we show additional results obtained by finetuning the Realistic Vision Model with FouRA adapters on our curated style datasets, \textit{3d}, \textit{Origami} and \textit{Paintings}. As observed, FouRA is capable of generating a diverse set of aesthetically appealing images.

\clearpage

\section{Additional Experiments for Text-to-Image Editing using Concept Sliders}
Concept sliders provide a framework to train LoRA adapters on single (image, prompt) pair (for example: "very old, wrinkly, gray hair, aged skin") in conjunction with multiple attributes (for example: Male person, very old etc). The disentanglement objective operates on the semantic space of diffusion models constraining the edit to occur only along the direction of the concept without changing the attributes. 

From \ref{fig:sv_distribution} we learnt that $\Delta W$ has a small eigen spread leading to more compact representation. Our method favous lower effective rank and the trained model naturally converges to decorrelated subspaces from the base model weights \ref{sec:decorr} . In addition in an informal proof \ref{sec:appendix_disentangled} we show that one can leverage the properties of FouRA to learn composition of concepts with less interference with the subspace of other concepts. 

We compare the performance of FouRA with LoRA when trained on explicit pairs of prompts across 20 different attributes acting as guidance. We train 3 sliders \textit{"curly hair"}, \textit{"surprise face"} and \textit{"Age slider"} on both the baseline LoRA and our adapter for upto 1000 steps. We trained the model on $rank = 8$.  We show that despite explicit training on pairs, low rank adapter space is still prone to changes in gender and race for strong adapter scales especially strength $\geq$ 4.  Below we show results on Single Adapter and Composite adapter. 

\begin{figure*}[h]
    \centering
    \includegraphics[width=0.7\linewidth]{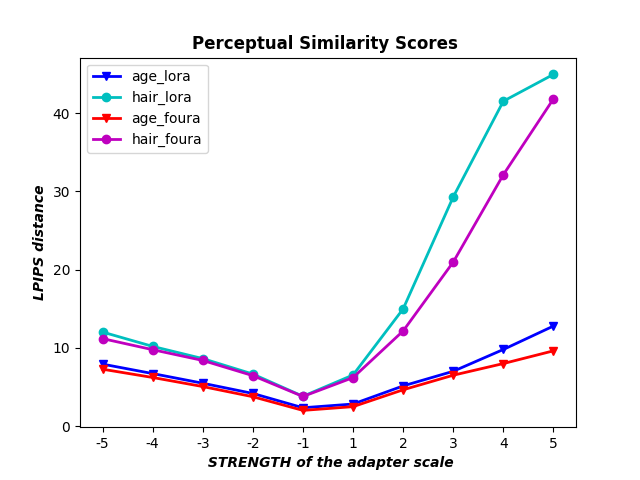}
    \caption{\textbf{Perceptual metric drops for LoRA compared to FouRA for the sliders on "age" and "hair"}. These were tested across 10 scales from (-5, 5). Similarity score was computed across 1000 images and 500 prompts of 10 seeds each. }
    \label{fig:lpips_sliders}

\end{figure*}

\paragraph{Single Concept}
We follow the SDEdit style inference where the adapter kicks in after $\mathcal{T} \in (750, 800, 850)$ timesteps. We notice that the effect of adapter in FouRA-DCT is far less below 800. Refer to figures below for more examples. For our results we fixed the $\mathcal{T} = 800$. We evaluate our results on LPIPS \ref{fig:lpips_sliders}.  While our adapter is far more stable compared to LoRA adapter between the strengths $[-6, 6]$. We also note that FouRA on DCT slightly better performance over FFT and for brevity we only show results on DCT. We note that FouRa maintains the balance between prompt and style fidelity and the quality of generated images. 

Below are some of the examples of \textit{Age}, 
\begin{figure*}[!h]
    \centering
    \includegraphics[width=1.0\linewidth]{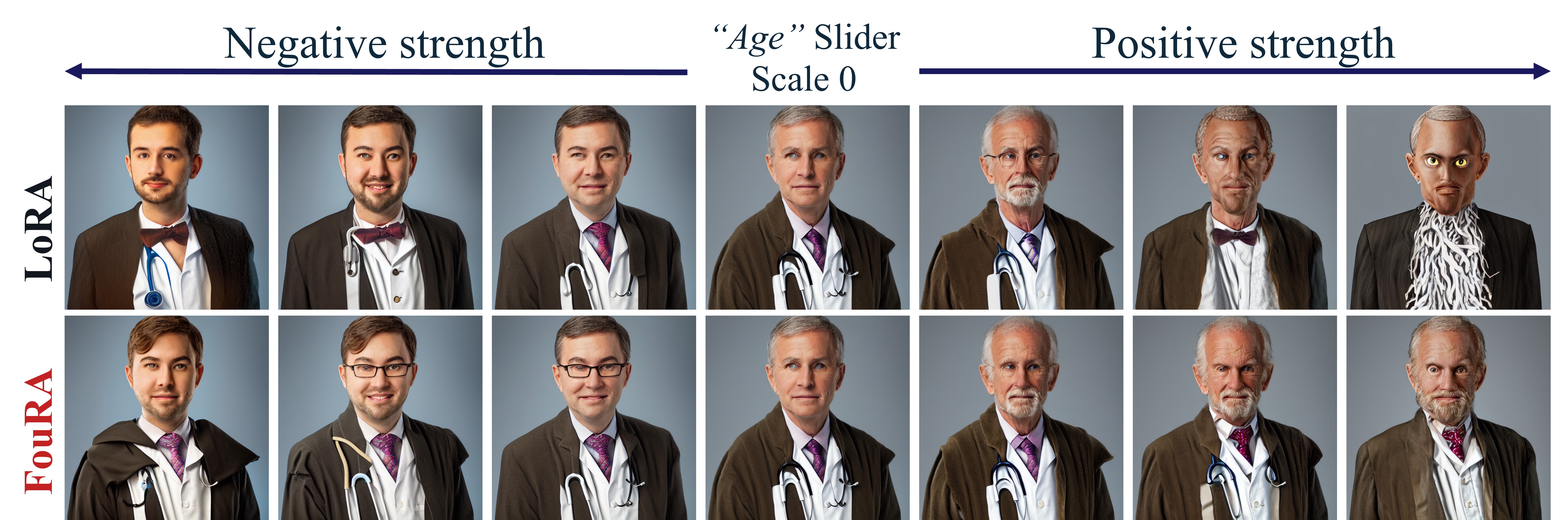}
    \caption{\textit{Age} Slider, LoRA (top) vs FouRA (bottom).  We find that as the strength increases there are more prominent skin tone variations in LoRA. }
    \label{fig:Age slider doctor}
\end{figure*}
\vspace{2.2 em}

\begin{figure*}[h]
    \centering
    \includegraphics[width=0.8\linewidth]{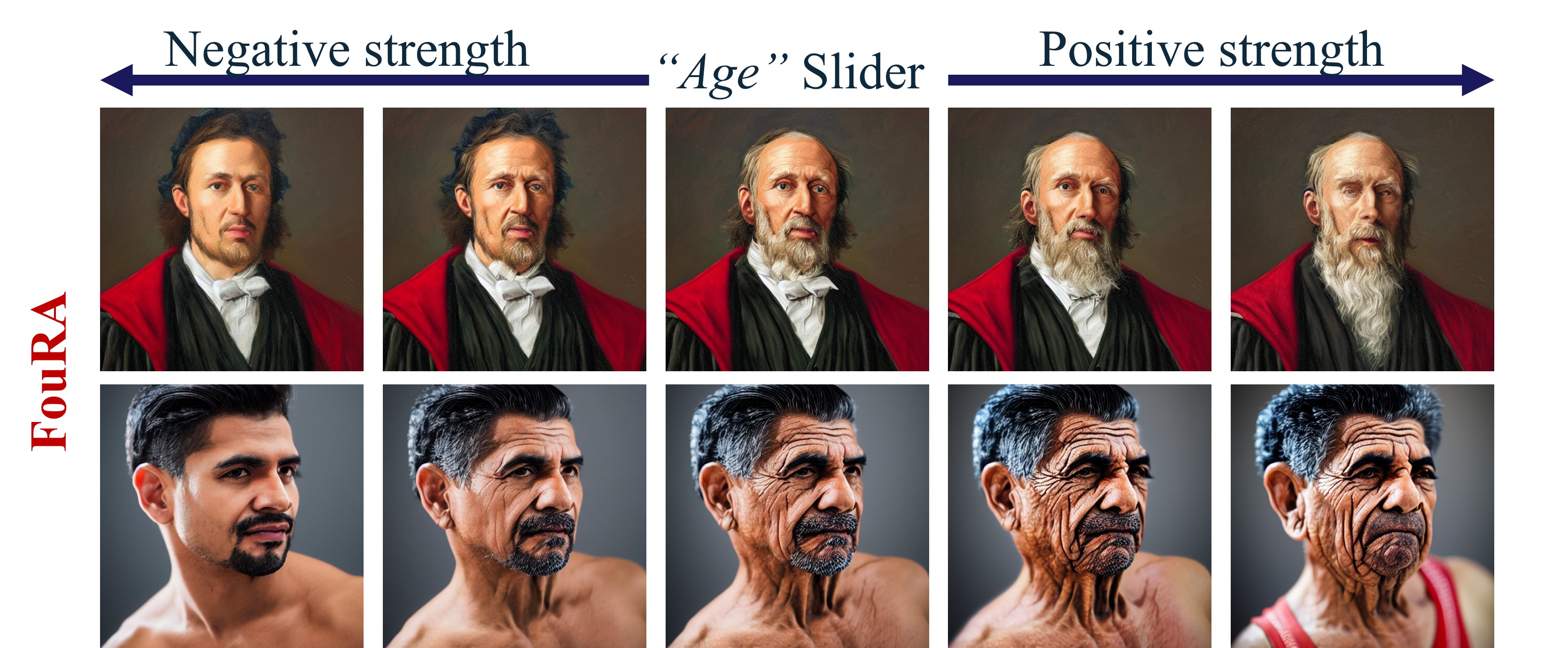}
    \caption{\textit{Age} FouRA Slider, "Portrait of a doctor" (top) and "Photo of an Hispanic man" (bottom).}
    \label{fig:Age slider other examples}
    
\end{figure*}

In general Age sliders shows a good improvement on LPIPS score for strength above 3 as shown in figure \ref{fig:lpips_sliders}. We notice that as the strength increases FouRA disentangles from other attributes better. 
 
We also train an adapter to change the strength of curls in hair. Below we show more examples for curly hair. We notice that the  both LoRA and FouRA adapters are  sensitive to increasing strength. As can be observed LPIPS score are higher for \textit{Hair} than for \textit{Age}.  As the strength increases the LoRA adapter tend move in the direction of increased prompt fidelity and removing the face of the person or crunching the face to add more details of hair in LoRA. We show the quanitative results for the same using LPIPS. We observe that across strengths $1 \leq 5$ the FouRA has much smaller LPIPS score. Please refer to the right figure in \ref{fig:age_slider_1}. Below we share more examples of FouRA on other prompts.  

\begin{figure*}[h]
    \centering
    \includegraphics[width=0.8\linewidth]{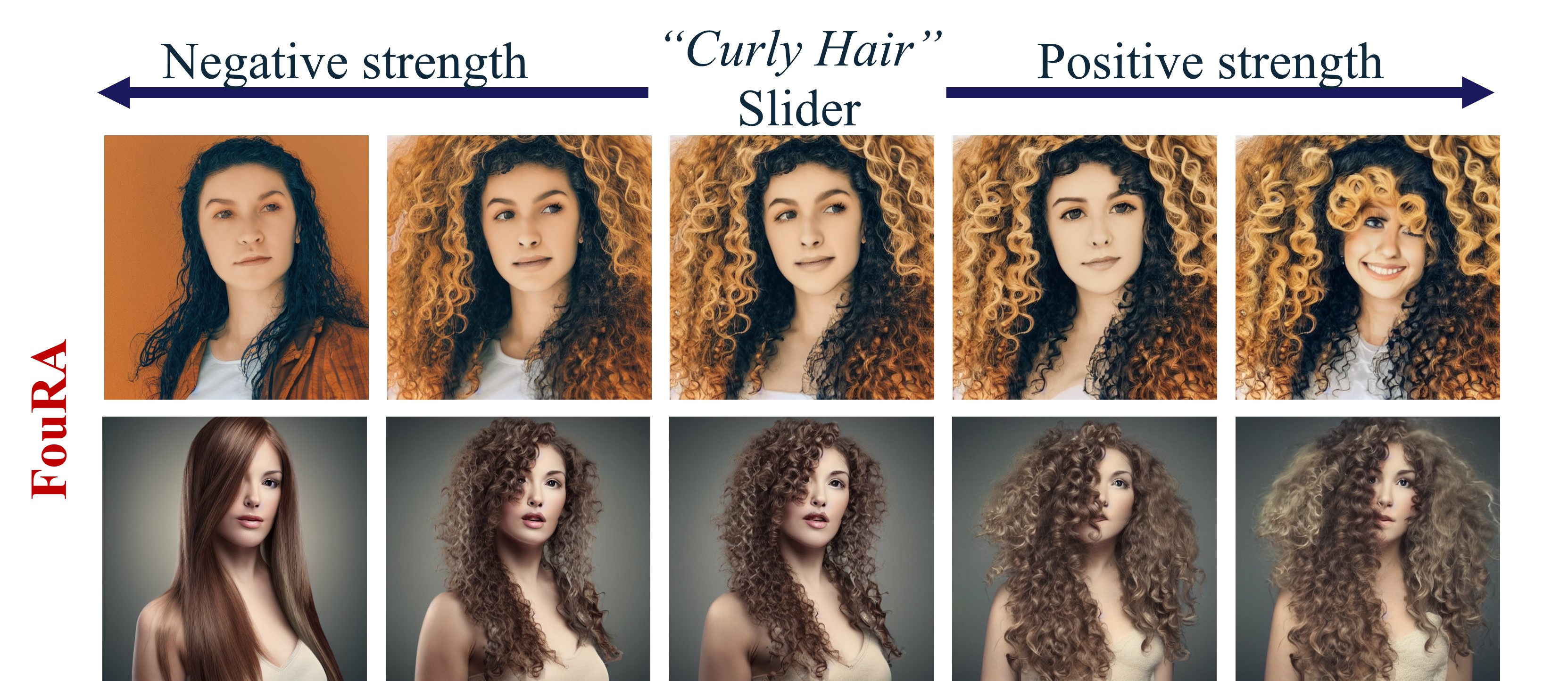}
    \caption{\textit{Hair} Slider: We find that as the strength of the adapter increases the curls increase. In the top image we also see minor variations in the facial details of the person. }
    \label{fig:Hair slider other examples}
\end{figure*}

\vspace{1.2 em}

\begin{figure*}[h]
    \centering
    \includegraphics[width=1.0\linewidth]{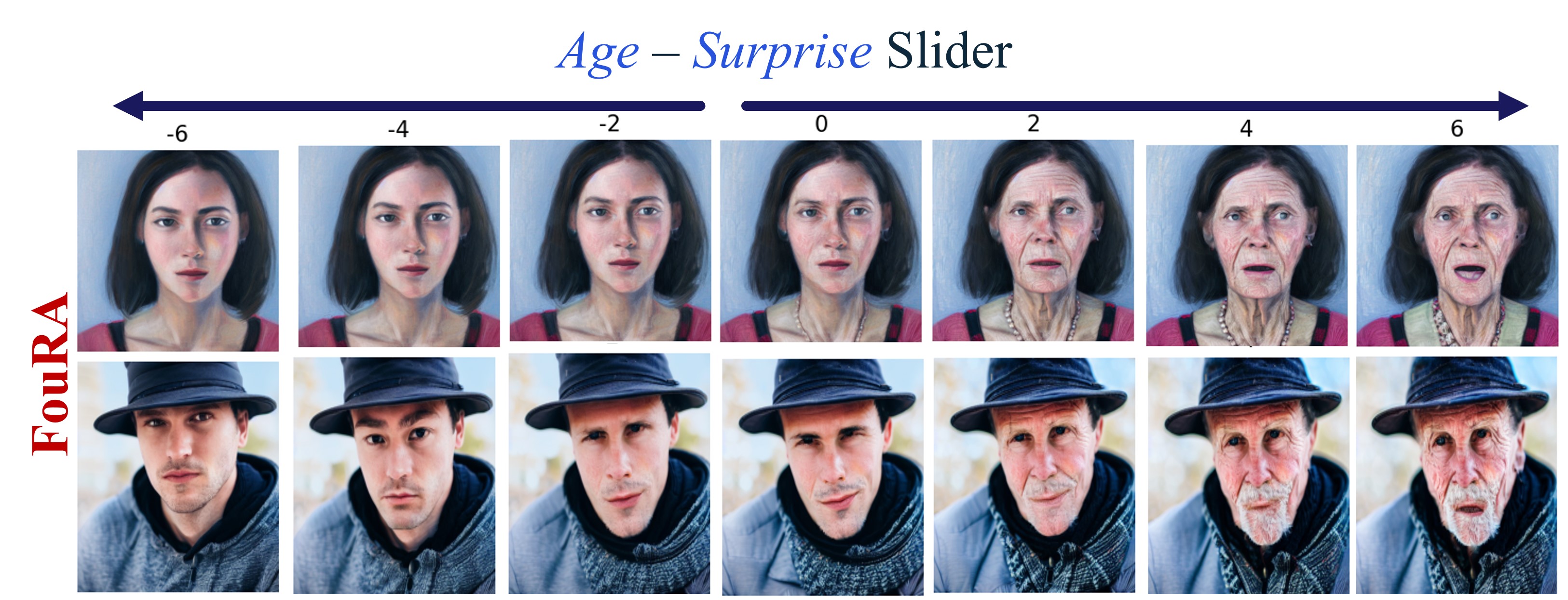}
    \caption{\textbf{Composite FouRA }. Composite \textit{surprised}, \textit{age} slider. Here we show the combined adapter as the strengths of each adapter are jointly incremented in each step in the image. The adapter strengths are (-6 6) for left most image and (6,6) for the right most image. The positive prompt for \textit{surprised face} prompt: \textbf{"looking surprised, wide eyes, open mouth"}}
    \label{fig:composite_slider_surprised}
    \vspace{- 1 em}
\end{figure*}

\paragraph{Composite LoRA}: Below we show the results for combining adapters. To combine adapters, we varied the strengths of Adapter 1 between $strengths \in (-8, 8) $ and Adapter 2 between $strengths \in (-8, 8) $. We show some examples of only FouRA \ref{fig:composite_slider_surprised} for combined \textit{hair} and \textit{Age} adapter. We show the images for when the adapter strengths are equal i.e increase from $(-6,6)$ to $(6,6)$.

Below we show comparison between LoRA and FouRA across different adapter strengths. We emphasize the effect when one slider for e.g \textit{"Age"} has a very high adapter strength  on the second slider when the strength is low (bottom left image). We observe that for LoRA the facial distortions when both adapter strengths are high (bottom right) are very evident. The Age adapter in general seems to interfere more with the  \textit{Hair} for higher strengths. 

\begin{figure*}[h]
    \centering
    \includegraphics[width=1.0\linewidth]{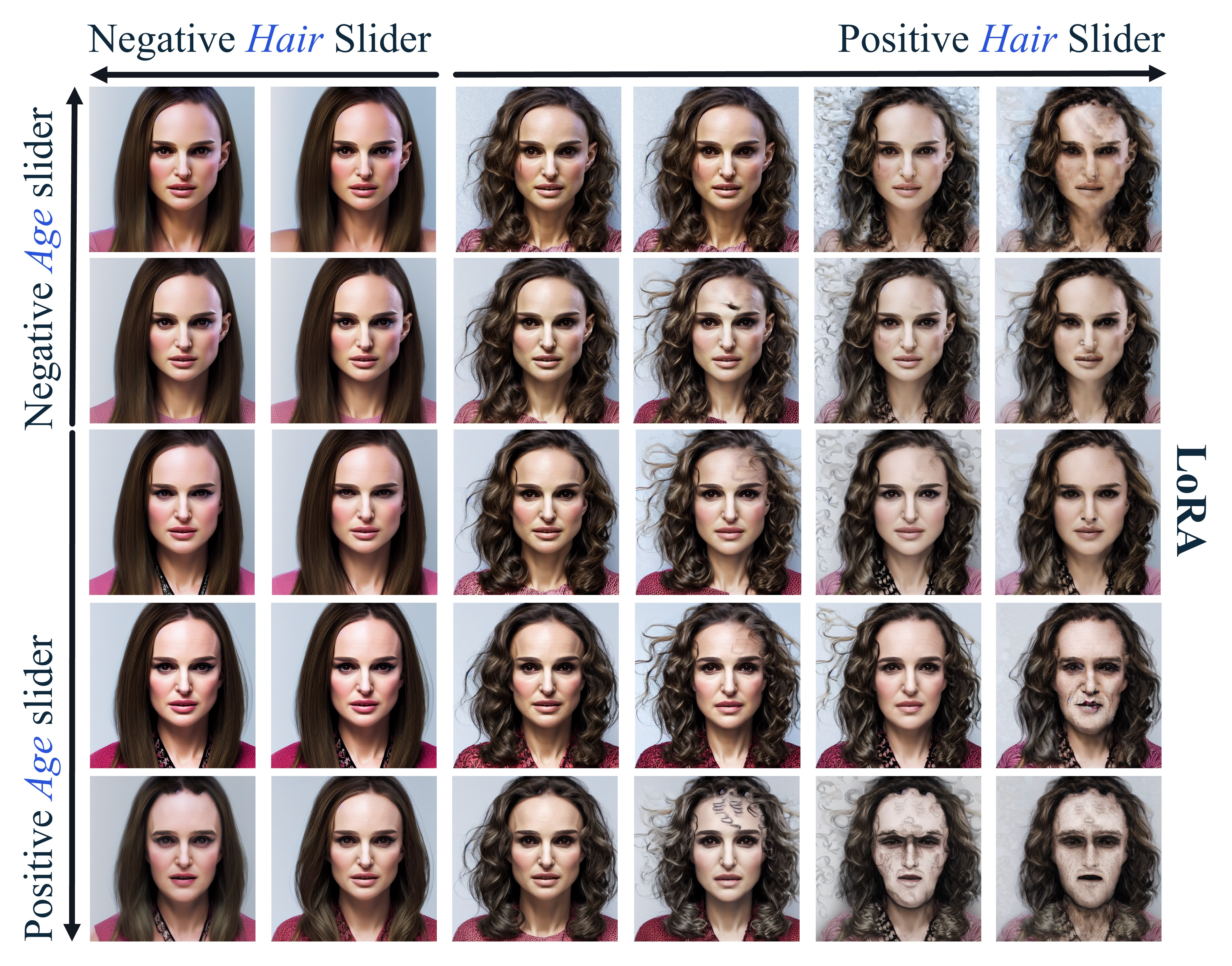}
    \caption{\textbf{{Composite LoRA} }. Composite \textit{hair}, \textit{age} slider. We find that for higher strength of \textit{Age} adapter as we increase the strength of \textit{Hair}, adapter seems to interfere with the facial features and almost distort the face. However for lower values of \textit{Hair} adapter. Here we show scales between {-6 to 8}}
    \label{fig:composite_slider_lora}
    \vspace{- 1 em}
\end{figure*}

\begin{figure*}[h]
    \centering
    \includegraphics[width=1.0\linewidth]{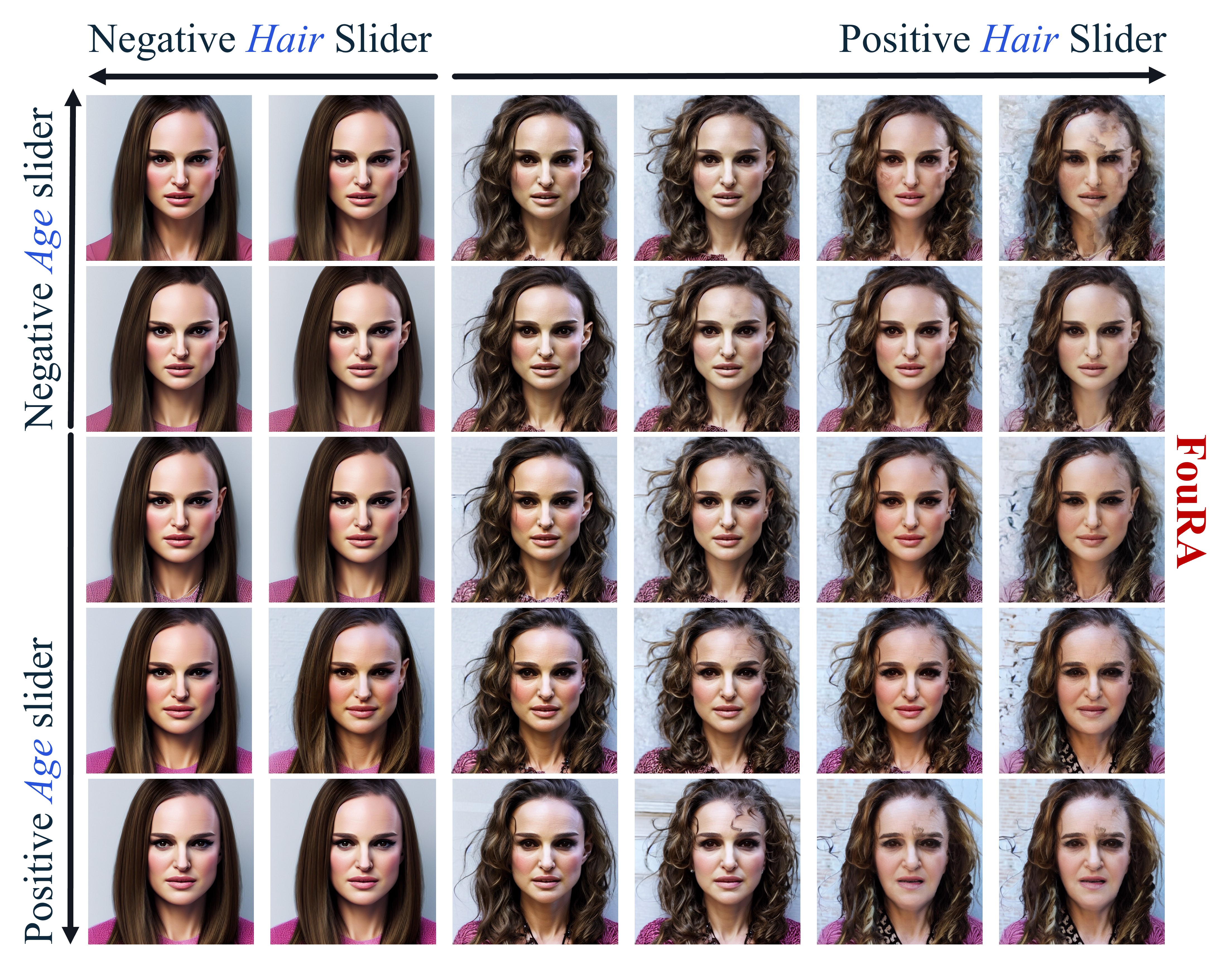}
    \caption{{\textbf{Composite FouRA }}. Composite \textit{hair}, \textit{age} slider. We note that the adapter is stable for many prompts and seeds upto scale of 8. There are artifacts at large scales strength upto scale=8 of positive slider, however we find that artifacts are fewer and don't distort the facial features. }
    \label{fig:composite_slider_foura}
    \vspace{- 1 em}
\end{figure*}

\section{Societal Impacts}
\label{sec:societal}
In this section, we discuss the societal impacts of our work. While there are benefits of training FouRA modules as highlighted in the main text, we consider that it can potentially have larger societal impacts. One of the major challenges of text-to-image models is digital forgery, highlighted in previous works~\cite{somepallidiffusion, somepalli2023understanding}. We observed that finetuning low-rank adapters on various tasks in image generation can lead to replication of the input image. This is due to the overfitting of LoRA on a small training set. However, we demonstrate in the paper how FouRA can push the generalization error bound further, hence resolving the data forgery problem to a great extent. Hence, we propose to utilize FouRA in applications where it is imperative to hide the training set, such that it can't be replicated.

\section{Limitations}
\label{sec:limitations}
FouRA, as demonstrated in the main text, is a highly effective parameter efficient fine-tuning method. However, as it makes use of frequency transforms (dft, dct), one potential limitation is that current Deep Learning hardware systems are not as optimal for frequency transform operations, as they are for matrix multiplies and convolutions. However, with astute recent works such as~\cite{si2023freeu, li2020fourier, nguyen2022transformer}, their popularity has increased in the field of Deep Learning. Hence, we foresee that it is only a matter of time before DL hardware systems get heavily optimized for frequency transforms.

\section{Future Work}
We have demonstrated that FouRA achieves great performance on tasks such as image generation, Image concept and style editing on Vision tasks in diffusion framework. A good extension of FouRA would be to explore the generalization capabilities to reuse the learnt basis on other adapters trained on different datasets. Additionally, for the FouRA module we would like to explore direct token masking in the frequency domain, as we observed some initial indicators, effectively correlating bands of frequencies and various characteristics of generated images. Finally, as we saw promising results on General Question Answering using RoBERTa, we would like to study FouRA's performance on multi-modal LLM's such as LLAMA-2B. Seeing the performance of FouRA, we feel encouraged to think that frequency domain fine-tuning of adapters will potentially be a popular research direction in the coming years.


\clearpage

\clearpage


\end{document}